\documentclass[10pt,twocolumn,letterpaper]{article}

\usepackage[pagenumbers]{cvpr} %

\usepackage{graphicx}
\usepackage{amsmath}
\usepackage{amssymb}
\usepackage{booktabs}

\usepackage[font=small,labelfont=bf]{caption}
\usepackage{color, colortbl}
\usepackage{xcolor}
\usepackage{amssymb}
\usepackage{enumitem}
\usepackage{pifont}%
\newcommand{\cmark}{\ding{51}}%
\newcommand{\xmark}{\ding{55}}%

\definecolor{citecolor}{HTML}{0071bc}
\usepackage[breaklinks=true,bookmarks=false,colorlinks,bookmarks=false, citecolor=citecolor, pagebackref]{hyperref}

\usepackage[toc,page,header]{appendix}
\usepackage{minitoc}
\usepackage[accsupp]{axessibility}

\usepackage[capitalize]{cleveref}
\crefname{section}{\S}{\S\S}
\crefname{subsection}{\S}{\S\S}
\Crefname{table}{Table}{Tables}
\crefname{figure}{Figure}{Figures}
\crefname{table}{Table}{Tabs.}

\newcommand{\benchmark}{GeneCIS\xspace}

\newcommand{\query}{$I^{R}$}
\newcommand{\target}{$I^{T}$}
\newcommand{\condition}{$c$}

\definecolor{Gray}{gray}{0.6}
\definecolor{babyblueeyes}{rgb}{0.63, 0.79, 0.95}
\definecolor{formalshade}{RGB}{213,230,242}

\definecolor{myblue}{RGB}{0, 77, 128}
\definecolor{myyellow}{RGB}{255, 217, 50}
\definecolor{mygreen}{RGB}{33, 140, 33}
\definecolor{myred}{RGB}{255, 100, 78}

\usepackage[strict]{changepage}

\usepackage{framed}

\newenvironment{formal}{%
  \MakeFramed{\advance\hsize-\width\FrameRestore}%
  \noindent\hspace{-4.55pt}%
  \begin{adjustwidth}{}{7pt}%
  \vspace{2pt}\vspace{-1pt}%
}
{%
  \vspace{-1pt}\end{adjustwidth}\endMakeFramed%
}

\def\cvprPaperID{629} %
\def\confName{CVPR}
\def\confYear{2023}

\begin{document}

\doparttoc %
\faketableofcontents %

\title{GeneCIS: A Benchmark for General Conditional Image Similarity}

\author{
Sagar Vaze$^{1,2}$\thanks{Work done during an internship at Meta AI Research.} \quad \quad
Nicolas Carion$^{1}$ \quad \quad
Ishan Misra$^{1}$ \\
$^{1}$ FAIR, Meta AI \quad \quad $^{2}$ VGG, University of Oxford\\
\small{Project Page:} \href{https://sgvaze.github.io/genecis/}{\texttt{sgvaze.github.io/genecis}}
}
\maketitle

\begin{abstract}

    We argue that there are many notions of `similarity' and that models, like humans, should be able to adapt to these dynamically.
    This contrasts with most representation learning methods, supervised or self-supervised, which learn a fixed embedding function and hence implicitly assume a single notion of similarity.
    For instance, models trained on ImageNet are biased towards object categories, while a user might prefer the model to focus on colors, textures or specific elements in the scene. 
    In this paper, we propose the GeneCIS (`genesis') benchmark, which measures models' ability to adapt to a range of similarity conditions.
    Extending prior work, our benchmark is designed for zero-shot evaluation only, and hence considers an open-set of similarity conditions.
    We find that baselines from powerful CLIP models struggle on GeneCIS and that performance on the benchmark is only weakly correlated with ImageNet accuracy, suggesting that simply scaling existing methods is not fruitful.
    We further propose a simple, scalable solution based on automatically mining information from existing image-caption datasets.
    We find our method offers a substantial boost over the baselines on GeneCIS, and further improves zero-shot performance on related image retrieval benchmarks. 
    In fact, though evaluated zero-shot, our model surpasses state-of-the-art supervised models on MIT-States.
\end{abstract}

\vspace{-8mm}
\begin{formal}
    \textit{We, the architects of the machine, must decide a-priori what constitutes its `world'; what things are to be taken as `similar' or `equal'} --- Karl Popper, 1963
    
\end{formal}
\vspace{-4mm}

\section{Introduction}
\label{sec:introduction}

\begin{figure}[!t]
    \centering
    \includegraphics[width=\linewidth]{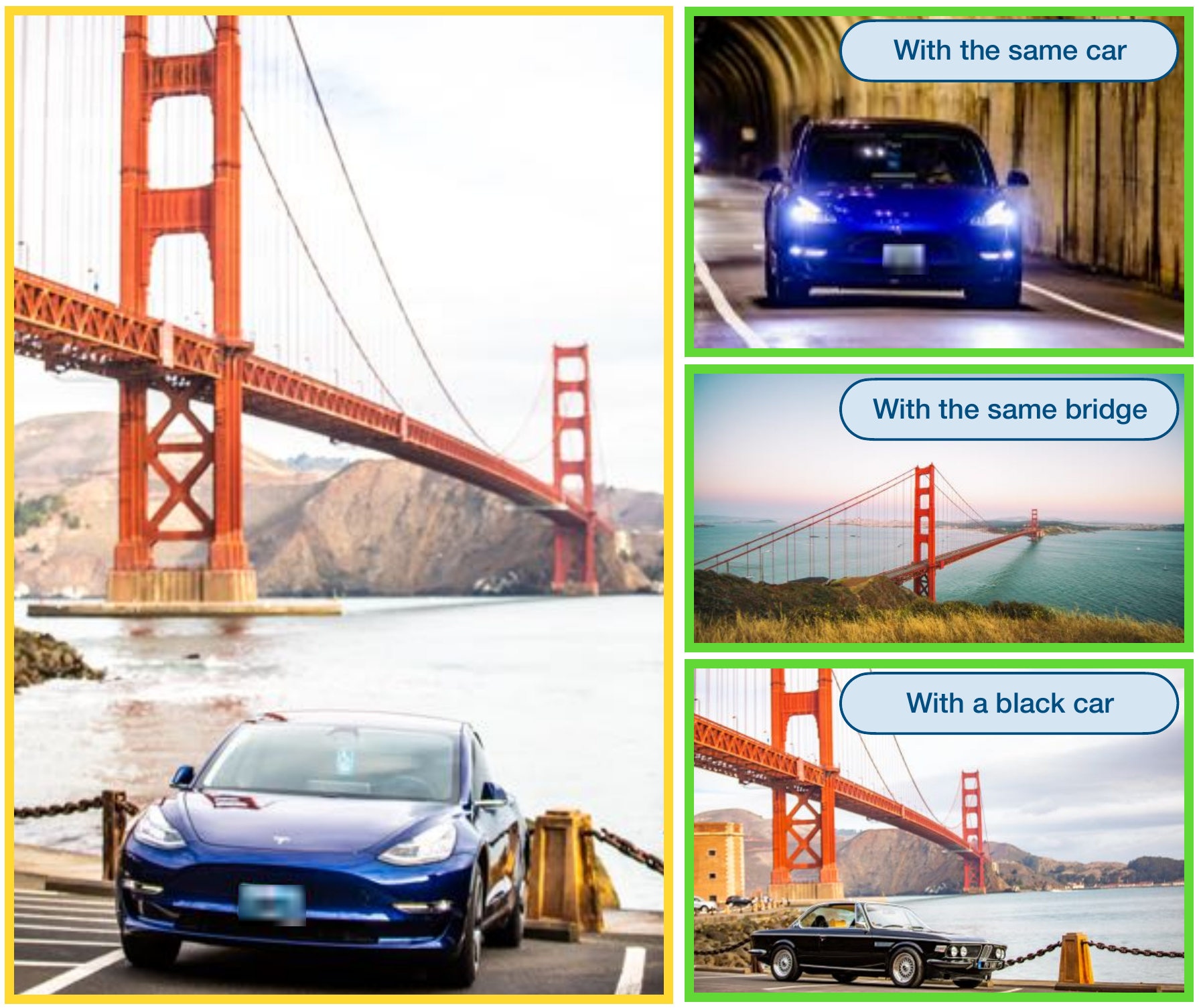}
    \caption{Given different \textit{conditions} (shown as blue text), different images on the right can be considered most `similar' to the reference on the left.
    We present a general way to train and evaluate models which can adapt to different notions of similarity.
    }
    \label{fig:motivation}
\end{figure}

Humans understand many notions of similarity and choose specific ones depending on the task at hand~\cite{goldstone2012similarity,popper1963}.
Consider the task of finding `similar' images illustrated in~\cref{fig:motivation}.
Which of the rightmost images should be considered `most similar' to the reference?
Given different \textit{conditions}, each image could be a valid answer. 
For instance, we may be interested in a specific object in the scene, focusing on either the `car' or `bridge'.
One could even indicate a `negative' similarity condition, specifying a \textit{change} in the image to identify the bottom image as most similar.

Learning such similarity functions is a central goal in discriminative deep learning \cite{chen2020improved,chen2020simple,Wang_2014_CVPR,roth2020revisiting,Kim_2020_CVPR,Taigman_2014_CVPR,chen2016face}.
Discriminative models, either supervised \cite{Wang_2014_CVPR,sym11091066} or self-supervised \cite{caron2021emerging,caron2020unsupervised}, learn embedding functions such that `similar' images are closer in feature space than `dissimilar' images.
However, since there are infinitely many notions of image similarity, how do we allow our models to choose?

Almost all current approaches assume a single notion of similarity, either by explicitly training on a specific concept \cite{Wang_2014_CVPR,Taigman_2014_CVPR} or through an implicit assumption in the underlying data distribution \cite{chen2020simple,caron2020unsupervised}.
Meanwhile, prior works tackling the conditional problem have focused on constrained domains such as fashion \cite{veit2017conditional,tanSimilarity2019} or birds \cite{mishraPAN2021}, with a restricted set of similarity conditions.
This is because developing and evaluating models that can adapt to generic notions of similarity is extremely challenging.
Specifically, curating data to train and evaluate such models is difficult, as collecting annotations for all concepts of similarity is impossible.

In this work we study the problem of general conditional image similarity, training on an open-set of similarity conditions, and evaluating on diverse similarity notions in a `zero-shot' manner.
We first design a benchmark comprising of \textit{four evaluation datasets} for conditional image similarity, setting up conditional retrieval tasks.
We define these tasks under a unified framework which spans practical use cases, and propose the benchmark as a sparse but broad coverage of the conditional similarity space.
We propose these datasets for \textit{zero-shot evaluation only},
and suggest that models which can perform well without fine-tuning can flexibly adapt to general notions of similarity, as desired.
We name this benchmark GeneCIS (`\textit{genesis}') for \underline{\textbf{Gene}}ral \underline{\textbf{C}}onditional \underline{\textbf{I}}mage \underline{\textbf{S}}imilarity.
On GeneCIS, we find that baselines built from powerful CLIP backbones struggle and, moreover, that performance on it is only weakly correlated with the backbones' ImageNet accuracy \cite{deng2009imagenet}.
This is in contrast to popular vision tasks such as segmentation \cite{li2022language} and detection \cite{minderer2022simple}, underlining the benchmark's utility. 

We also propose a solution to training general conditional similarity models, based on parsing large-scale caption datasets \cite{sharma2018conceptual,schuhmann2022laion}. 
Rather than requiring exhaustive similarity annotations, we find that we can automatically mine this information from already abundant image-caption data. 
We show that training in this way offers substantial gains over the baselines, approaching (and in some cases surpassing) carefully designed specific solutions for each of the GeneCIS tasks.
In addition, we demonstrate that our method scales with increasing amounts of caption data, suggesting promising directions for future work.
Finally, on related benchmarks from the `Composed Image Retrieval' (CIR) field \cite{vo2019composing,liu2021cirplant}, we find our method provides gains over zero-shot baselines.
In fact, our model outperforms state-of-the-art on the MIT-States benchmark \cite{StatesAndTransformations}, despite being evaluated zero-shot and never seeing the training data.

\par \noindent \textbf{Contributions.} 
\textcolor{mygreen}{\textbf{(i)}} We present a framework for considering conditional image similarity, an important but understudied problem; 
\textcolor{mygreen}{\textbf{(ii)}} We propose the GeneCIS benchmark to test models' abilities to dynamically adapt to different notions of similarity; 
\textcolor{mygreen}{\textbf{(iii)}} We show that current vision-language models like CLIP struggle on GeneCIS, and that performance on it is only weakly correlated with ImageNet accuracy;
\textcolor{mygreen}{\textbf{(iv)}} We design a scalable solution to the conditional similarity problem based on automatically parsing large-scale image-caption data;
\textcolor{mygreen}{\textbf{(v)}} We show our models provide substantial gains over zero-shot CLIP baselines;
\textcolor{mygreen}{\textbf{(vi)}} We validate our models on related CIR benchmarks,
surpassing state-of-the-art on MIT-States despite zero-shot evaluation.
\begin{figure*}[ht!]
    \centering
    \setlength\belowcaptionskip{-13pt}
    \setlength\abovecaptionskip{2pt}
    \includegraphics[width=\linewidth]{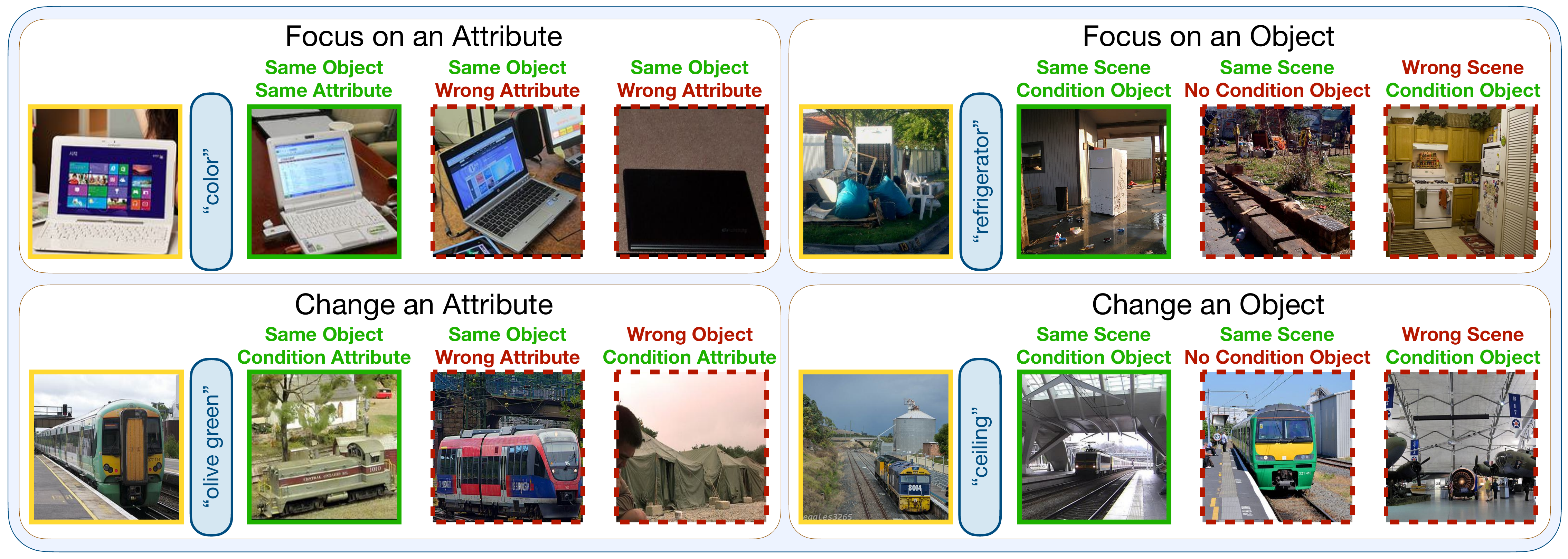}
    \caption{\textbf{The GeneCIS benchmark} contains four evaluation tasks for conditional similarity, where the goal is to retrieve the most similar image from a gallery 
    (right, green squares), given a reference (left, yellow squares), and condition (blue ovals).
    Each task explores one combination of `focus'/`change' an `attribute'/`object'.
    All galleries contain `distractors' (dashed, dark-red squares) which are \textit{implicitly} similar to the reference or condition. 
    Thus, given a reference and explicit condition, GeneCIS evaluates models' ability to select the \textit{most conditionally similar} gallery image.
    Note: We show three gallery images for clarity, though all GeneCIS galleries have 10-15 images. 
    }
    \label{fig:setting}
\end{figure*}

\section{Related Work}
\label{sec:related}

Our thesis that the similarity between two images should be conditional is generally relevant to the \textit{representation learning} literature, which aims to learn embedding functions based on a single (often implicit) notion of similarity.

For instance, \textit{deep metric learning} \cite{roth2020revisiting,sym11091066,Kim_2020_CVPR} aims to learn visual representations such that images from the same category are projected nearby in feature space.  
This idea is used in practical domains such as \textit{image retrieval} \cite{Brown2020smoothap, Revaud_2019_ICCV, Radenović_2018_CVPR}, \textit{face verification} \cite{NIPS2014_e5e63da7,chen2016face,Taigman_2014_CVPR} and \textit{vehicle re-identification} \cite{liu2016reid, KHAN201950, He_2019_CVPR}.
The key limitation here is that networks are trained to encode a single notion of similarity, namely category-level similarity.
While some work considered notions of similarity at different visual granularities \cite{touvron2021grafit,cui2019measuring,berman2019multigrain}, we posit that there exist concepts of similarity (\eg shape and color) which are orthogonal to categories.

Meanwhile, \textit{contrastive learning} \cite{chen2020simple,chen2020improved,caron2020unsupervised,caron2021emerging} defines notions of similarity by specifying a set of transformations to which the representation should be invariant (\eg color jitter or random cropping), encouraging augmentations of the same instance to be embedded together.
Similarly, \textit{vision-language} contrastive training \cite{Radford2021Learning,jia2021scaling} learns joint embedding spaces, where 
images' representations are aligned with their paired captions. 
Though the precise notions of similarity are difficult to define in this case, we note that the embeddings are fundamentally unconditional, with a single deterministic embedding of a given image.

Finally, we highlight three relevant sub-fields in the literature: \textit{conditional similarity networks} (CSNs); \textit{compositional learning} (CL); and \textit{composed image retrieval} (CIR).
CSNs are networks with multiple subspaces for different notions of similarity \cite{veit2017conditional}.
Though their motivation is highly related to our work, CSNs are trained in a supervised manner with pre-defined similarity conditions \cite{veit2017conditional,mishraPAN2021,Lin_2020_CVPR}, 
and/or are evaluated in constrained domains such as fashion \cite{tanSimilarity2019,kimMTLearning2021}.
In contrast, we aim to train on an open-set of similarity conditions and evaluate zero-shot on natural images.
Meanwhile, our work is related to CL research in that we seek to compose information from images and conditions to establish similarities. 
However, again, CL models are often assessed on their ability to recognize unseen combinations of a finite set of visual primitives \cite{misra2017red,Pham_2021_CVPR,perez2018film}.
Lastly, the most similar setup to GeneCIS is proposed in the recent CIR~\cite{vo2019composing}.
It tackles the problem of composing an image and text prompt to retrieve relevant images from a gallery \cite{delmas2022artemis,baldrati2022conditioned,Anwaar_2021_WACV}. 
This is typically posed in the context of fashion \cite{guo2019fashion,han2017automatic}, with the text prompt acting as an image edit instruction (\eg `the same dress but in white' \cite{Anwaar_2021_WACV}). 
As such, CIR tackles a subset of the conditional similarity problem, by presenting models with a `negative' similarity condition.

\par \noindent \textbf{Key similarities and differences with prior work:} In this work, we leverage CIRR \cite{liu2021cirplant} and MIT-States \cite{StatesAndTransformations} (natural image CIR datasets) for additional evaluations, and further leverage the `Combiner' architecture \cite{baldrati2022conditioned} to compose text conditions and image features.
Broadly speaking, our work differs from CSNs, CL and CIR in that we do not train on a finite, closed-set of similarity conditions or visual primitives.
Instead, we train models on open-world image-caption data, and demonstrate a flexible understanding of conditional similarity through zero-shot evaluation on a range of similarity conditions in natural images.

\section{Conditional Similarity}
\label{sec:problem_definition}

We now describe our setup for the conditional similarity problem and its associated challenges -- both with benchmarking models and acquiring data to train them.
In~\cref{sec:benchmark_details} we introduce the \benchmark benchmark which measures important aspects of the problem.
In~\cref{sec:method}, we present a scalable solution to automatically acquire training data from widely available image-caption datasets.

\par \noindent \textbf{Problem Definition:}
We define the problem of conditional similarity as learning a similarity function between two images given an \emph{explicit} condition:
$f(I^{T}; I^{R}, c)$
yields the scalar similarity between a target image, 
\target
, and a reference image,
\query,
given some external condition,
\condition.
We use the scalar $f(\cdot)$ to find the most conditionally similar image from a target set, \ie, to solve a retrieval task.
In this work we consider the condition to be a user-specified text prompt, although other types of condition are possible.
We highlight that standard image similarity, framed as $f(I^{T}, I^{R})$, \emph{implicitly} assumes a similarity condition, often incorporated into the model or dataset (see ~\cref{sec:related}).
We refer to the case where images are similar under an unspecified condition as the images being \emph{implicitly similar}.

\subsection{Challenges in training and evaluation}
\label{sec:condsim_challenges}

\par \noindent \textbf{Challenges in evaluation:}
The key difficulty in evaluating conditional similarity is that there are infinitely many possible conditions: from `images with the same top-left pixel value are similar' to `the same image but upside down is similar'.
Thus, it is impossible to evaluate models' ability to adapt to \emph{every} similarity condition.
Instead, in~\cref{sec:benchmark_details}, we introduce the \benchmark benchmark which consists of a subset of such conditions, and covers a broad range of practical use cases.
We suggest that models which produce \textit{zero-shot} gains across GeneCIS, without finetuning, are more capable of flexibly adapting to different notions of similarity.

\par \noindent \textbf{Challenges in acquiring training data:}
Since the space and diversity of similarity conditions is huge, acquiring human annotations to train for \emph{every} type of conditional similarity is not feasible.
For instance, to train a function which is sensitive to object category given some conditions (\eg, `car' or `bridge' objects in ~\cref{fig:motivation}), and `color' given others (\eg `blue' or `black' car in ~\cref{fig:motivation}), we need training data containing both features.
Prior work addresses this by dramatically restricting the space of conditions and training on human annotations for pre-defined notions of similarity \cite{veit2017conditional, mishraPAN2021}.
In~\cref{sec:method}, we describe an automatic method which leverages existing large-scale image-text datasets to learn an open-set of similarity conditions.
The resulting model can be evaluated in a zero-shot manner across different types of conditional similarity task.

\section{The \benchmark Benchmark}
\label{sec:benchmark_details}

GeneCIS considers two important dimensions of the conditional similarity problem.
Firstly, a user may be interested in an \textit{object} in the scene (`with the same \underline{car}') or an \textit{attribute} of a given object (`the same \underline{color} as the car'). 
Secondly, the condition could either \textit{focus} on a particular aspect of the image (`the \underline{same} color as the car') or specify the `negative' space of a similarity condition,
by defining a \textit{change} in the image (`this car \underline{but in black}').

We propose \textbf{four evaluation tasks in GeneCIS}, that covers the combination of the above dimensions and hence a diverse range of conditional similarities.
For each of the tasks, we construct retrieval problems with: a reference image, \query; a text condition, \condition; and a retrieval gallery of $M$ target images, 
$\{I^{T}_{i}\}_{i=1}^{M}$, of which only one is `correct' or `positive'.
The task is to identify which of the target images is most similar to the reference, given the condition.
The retrieval tasks, illustrated in~\cref{fig:setting} with more examples in ~\cref{sec:supp_genecis_examples}, are:

\vspace{-2mm}
\begin{itemize}[leftmargin=*]
\itemsep0.1em 
    \item \textbf{Focus on an Attribute:} This task evaluates a model's ability to focus on a specific attribute type (e.g `color' or `material').
    For instance, in ~\cref{fig:setting}, we see a white laptop and the condition `color', with the task being to select the laptop with the same color from the gallery.
    \item \textbf{Change an Attribute:} This task contains `negative' similarity conditions, considering target images with a specific attribute changed to be most similar.
    In ~\cref{fig:setting}, the aim is to retrieve the same object (`train') but with the color changed from `green' to `olive green'.
    \item \textbf{Focus on an Object:} This task considers reference images with many objects, and we refer to the set of objects together as a proxy for the image `scene'. 
    The condition selects a single object from the reference as the most important (\eg `refrigerator' in ~\cref{fig:setting}) and the `positive' target contains the condition object as well as the same `scene' (\eg also contains `sky', `chair' \etc in ~\cref{fig:setting}).
    \item \textbf{Change an Object:} This task considers `negative' similarity through conditions which specify an object to be added to a scene.
    For instance, in ~\cref{fig:setting}, `ceiling' is specified, with the aim being to retrieve the same scene (a train station) but with a ceiling also present.
\vspace{-2mm}
\end{itemize}

The tasks in GeneCIS are designed to be diverse and challenging for a single model while remaining well-posed.
In ~\cref{fig:setting}, given only the reference image, \query, and text condition, \condition, a human can readily identify which of the target images is most `similar'. 
We wish to benchmark vision models' competency at the same task.

For the benchmark to be challenging, we would want the model to need both the image content and the text condition to solve the problem.
Thus, we include different forms of `distractor' images in the galleries.
For instance, for tasks with objects in the condition, we include distractors which have a similar `scene' to the reference but do not contain the condition object.
Such distractors are likely to affect models which are over-reliant on information from the reference image, without considering the condition.
Similarly, we include distractors which contain the object specified in the condition, but not the reference scene, confusing models which solely rely on the condition.
Meanwhile, for the attribute-based tasks, we include distractors which contain the reference object category, but not the correct attribute, and vice-versa.
As such, many targets are \textit{implicitly similar} to the reference (similar given some condition), but the positive image is the most similar \textit{given the provided condition}.

\begin{table}[t]
\footnotesize
\centering
\setlength\belowcaptionskip{0pt}
\caption{\textbf{Statistics} of the four tasks in the GeneCIS benchmark.}
\label{tab:benchmark_statistics}
\resizebox{\linewidth}{!}{
\begin{tabular}{llcc}
\toprule
Name                  & Base Dataset & \# Templates & \# Gallery Images \\
\midrule
Focus on an Attribute        & VAW \cite{Pham_2021_CVPR} & 2000       & 10            \\
Change an Attribute        & VAW \cite{Pham_2021_CVPR} & 2112       & 15            \\
Focus on an Object        & COCO \cite{coco2017panoptic} & 1960       & 15            \\
Change an Object        & COCO \cite{coco2017panoptic} & 1960       & 15            \\
\bottomrule
\end{tabular}
}
\end{table}
\begin{figure}[t]
    \centering
    \setlength\belowcaptionskip{-13pt}
    \setlength\abovecaptionskip{3pt}
    \includegraphics[width=\linewidth]{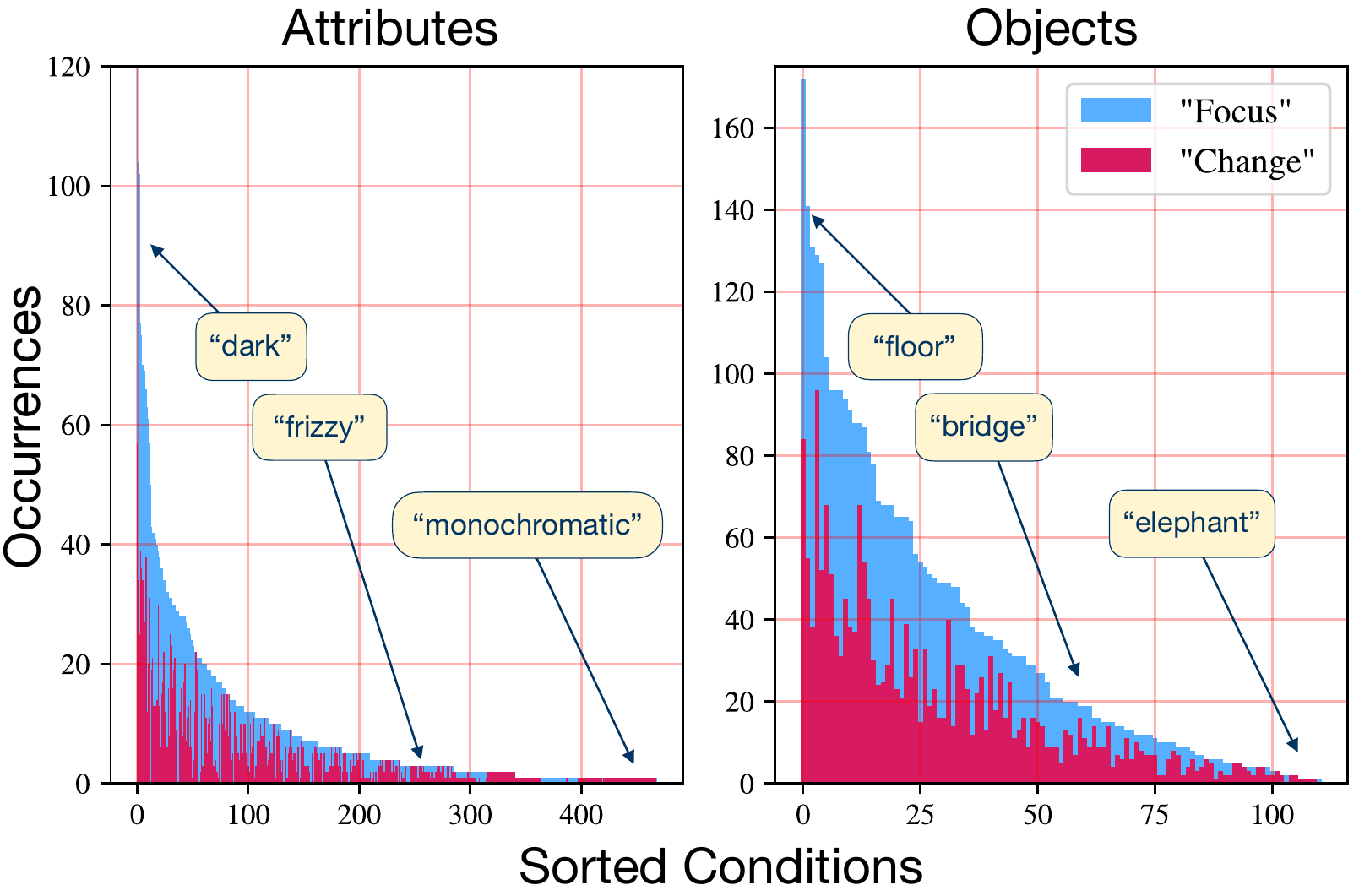}
    \caption{\textbf{Distribution of conditions} for attribute- and object-based conditions.
    For `Focus on an Attribute', we show the distribution of the common attribute between the reference and positive target image (the condition itself is an attribute type, \eg `color').}
    \label{fig:condition_distributions}
\end{figure}
\par \noindent \textbf{Benchmark Details:}
We construct all tasks by re-purposing existing public datasets.
For the two tasks which `focus on' and `change' \textit{attributes}, we leverage the VAW dataset~\cite{Pham_2021_CVPR}, which inherits from Visual Genome \cite{krishnavisualgenome}.
From VAW, we extract crops for individual objects and, for each object, use annotations for: object category; positively labelled attributes (which the object definitely possesses); and negatively labelled attributes (which the object definitely does not possess).
For the two tasks which `focus on' or `change' \textit{objects}, we use 
COCO Panoptic Segmentation data \cite{coco2017panoptic,lin2014coco}
containing dense category annotations for every pixel in the image.
We give full details of the template construction process for each task in \cref{sec:supp_task_construction}. 

\begin{figure*}[!t]
    \centering
    \setlength\belowcaptionskip{-15pt}
    \setlength\abovecaptionskip{2pt}
    \includegraphics[width=\linewidth]{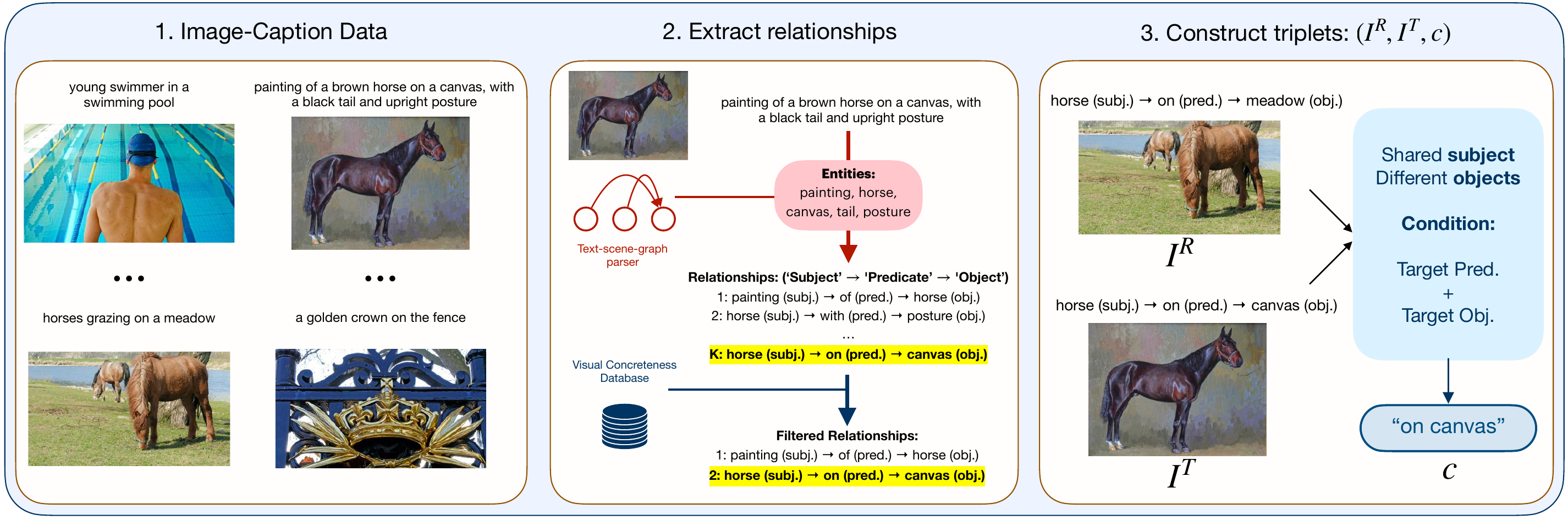}
    \caption{\textbf{Method overview.} Our method for training general conditional similarity functions extracts information from large-scale image-caption datasets (left). 
    We extract `Subject' $\rightarrow$ `Predicate' $\rightarrow$ `Object' relationships from the caption data (middle), before using them to construct training triplets where a \textit{reference} and \textit{target} image are related by a \textit{condition} (right).
    }
    \label{fig:method}
\end{figure*}

We show statistics of the evaluations in ~\cref{tab:benchmark_statistics}, including the number of retrieval templates and number of gallery images.
We note that we carefully construct the benchmarks such that there is only one `positive' image among the targets, with gallery sizes of between 10 and 15 images.
This is different to many `text-to-image' or `image-to-text' retrieval benchmarks \cite{lin2014coco,flickrentitiesijcv}, which contain galleries with thousands of targets.
Though larger galleries increase the tasks' difficulty,
the galleries inevitably contain some valid targets which are treated as negative.  
We further show the distribution of objects and attributes specified in the conditions in ~\cref{fig:condition_distributions}, noting that our space of conditions spans a long-tail of over 400 attributes and 100 objects.

\par \noindent \textbf{Noise and human verification:}
Though, in principle, our benchmark should be error free, manual inspection of the templates shows that noise is introduced through underlying inconsistencies in Visual Genome \cite{krishnavisualgenome}, VAW \cite{Pham_2021_CVPR} and COCO \cite{coco2017panoptic}.
We are currently in the process of collecting manual annotations and human verification of the templates, and present the current version as `GeneCIS v0'.

\section{Method}
\label{sec:method}

In ~\cref{sec:preliminaries}, we briefly describe preliminaries for our approach to learning general conditional similarity functions. 
This includes the model architecture and optimization objective which we inherit from prior work \cite{baldrati2022conditioned}.
In ~\cref{sec:triplet_mining}, we describe our main methodological contribution: an automatic and scalable way of mining conditional similarity training data from widely available image-caption datasets.

\subsection{Preliminaries}
\label{sec:preliminaries}

\par \noindent \textbf{Training data.}
To learn a conditional similarity function $f(\cdot)$, we train with triplets $(I^{R}, I^{T}, c)$,
where \query and \target are termed reference and target images, 
and $c$ is the condition defining a relationship between them.

\par \noindent \textbf{Model Architecture}
We parametrize the conditional similarity function $f(\cdot)$ with deep networks, first encoding features for 
$($\query, \target, \condition$)$
as 
$(\mathbf{x}^{R}, \mathbf{x}^{T}, \mathbf{e}) \in \mathbb{R}^{D}$.
We learn separate encoders, $\Phi(I)$ and $\Psi(c)$, for the images and text condition.
Next, we train a `Combiner' network \cite{baldrati2022conditioned}, which composes the reference image features with the condition text features as
$g(\mathbf{x}^{R}, \mathbf{e}) \in \mathbb{R}^{D}$.
Finally, we consider the scalar conditional similarity to be the dot product between the combined feature, and target image feature, as:
$f(I^{T}; I^{R}, c) = g(\mathbf{x}^{R}, \mathbf{e}) \cdot \mathbf{x}^{T}$.
Details of the Combiner architecture can be found in \cref{sec:supp_architecture} and ~\cite{baldrati2022conditioned}.

We initialize our image and text backbones, $\Phi(\cdot)$ and $\Psi(\cdot)$, with CLIP \cite{Radford2021Learning}. 
CLIP models are pre-trained on 400M image-text pairs containing a range of visual concepts.
Furthermore, the visual and text embeddings from CLIP are aligned, making it easier to learn the composition between reference image and conditioning text features.

\par \noindent \textbf{Optimisation Objective}
Given a batch of triplets,
$B = \{(I^{R}_i, I^{T}_i, c_i)\}_{i=1}^{|B|}$, we get features as
$\{(\mathbf{x}^{R}_{i}, \mathbf{x}^{T}_{i}, \mathbf{e}_{i})\}_{i=1}^{|B|}$.
Then, given a temperature $\tau$, we optimise $(\Phi, \Psi, g)$ with a contrastive loss \cite{oord2018representation}, as: 
\vspace{-5mm}

\begin{equation}
\label{eq:contrastive}
\medmuskip=2mu
\thickmuskip=3mu
\renewcommand\arraystretch{1.5}
\mathcal{L}=
- \frac{1}{|B|}  \sum_{i \in B}
\log \frac
{\exp \left(g(\mathbf{x}^{R}_i, \mathbf{e}_i) \cdot \mathbf{x}^{T}_i / \tau\right)}
{\sum_{j \in B} \exp \left(g(\mathbf{x}^{R}_i, \mathbf{e}_i) \cdot \mathbf{x}^{T}_j / \tau\right)}
\end{equation}

\subsection{Scalable training for conditional similarity}
\label{sec:triplet_mining}

To train for general conditional similarity, we wish to curate triplets for training,
$\mathcal{D}_{train} = \{(I^{R}_i, I^{T}_i, c_i)\}_{i=1}^N$,
with diverse conditions and concepts of similarity.
However, as the space of conditions increases, the burden for exhaustively annotating such a dataset increases exponentially. 
Instead, our method (illustrated in ~\cref{fig:method}) automatically mines training triplets from existing data sources:

\par \noindent \textbf{Image-caption Data:} We begin with large-scale image-caption data scraped from the internet, containing images paired with descriptive captions \cite{sharma2018conceptual, ordonez2011sbu}. 
We hope that the captions contain information about the objects and attributes in the image, which we can utilize for the conditional similarity task.
We also hope that such a method can scale with increasing data in the same way that conventional representation learning algorithms do. 

\par \noindent \textbf{Extract relationships:} We use an off-the-shelf text-to-scene-graph parser \cite{wu2019parser, schuster2015generating} to identify `Subject' $\rightarrow$ `Predicate' $\rightarrow$ `Object' relationships within the caption \cite{peyre2019detecting}.
For instance, from the central image in ~\cref{fig:method}, we extract the highlighted relationship `Horse' $\rightarrow$ `on' $\rightarrow$ `Canvas'.
Note that one caption may contain many such relationships.

We find that many of the entities (`Subjects' or `Objects') extracted by the parser are not visually grounded in the image, \eg, pronouns (`I', `you') or time-based nouns (`today', `yesterday').
To address this, we introduce an additional filtering step, where every entity is scored for `visual concreteness' based on a pre-existing database \cite{brysbaert2014lemmas}.
The database contains human ratings between 1 and 5 for how visually apparent a noun is.
For each extracted relationship, we average its `Subject' and `Object' concreteness scores, discarding relationships if their value is below a threshold.

\par \noindent \textbf{Construct triplets:} 
We first randomly select a relationship, taking the image it comes from as the `reference', \query. 
Having identified the \textit{subject} of the relationship (\eg `Horse' in the rightmost column of ~\cref{fig:method})
we identify all other relationships in the dataset containing the same subject. 
From this restricted pool of relationships, we randomly sample a `target' relationship and image, \target, with the same subject but a different \textit{object} (\eg a horse on a `canvas' instead of in a `meadow' in ~\cref{fig:method}).
Finally, we define the \textit{condition} of the triplet, \condition, as the concatenated `Predicate' and `Object' from the target relationship (`on canvas' in ~\cref{fig:method}).

\par \noindent \textbf{Discussion:}
We note that our mined triplets exhibit a bias towards the `Change an Object' GeneCIS task.
However, the triplets often involve abstract relationships between reference and target images (\eg `Horse on canvas' in ~\cref{fig:method}).
As such, solving the training task requires the model to use the condition to extract and modify diverse forms of information from the reference, which is the central requirement of the broader conditional similarity problem.

\section{Main Experiments}
\begin{table*}[ht]
\footnotesize
\centering
\caption{\textbf{Evaluation on \benchmark.} We evaluate baselines and our method.
We also evaluate specific solutions for each task (shown gray, these are not general conditional similarity functions and hence cannot be evaluated on all tasks).
Both across ten random seeds, and with ten cross-validation splits, we find a standard deviation of $\mathbf{\approx0.2\%}$ in our model's R@1 on each task, as well as on average over all tasks.
}
\label{tab:main}
\resizebox{\linewidth}{!}{
\begin{tabular}{l|ccc|ccc|ccc|ccc|cc}
\toprule
& \multicolumn{3}{c}{Focus Attribute} & \multicolumn{3}{c}{Change Attribute} & \multicolumn{3}{c}{Focus Object} & \multicolumn{3}{c}{Change Object} \\
\cmidrule(rl){2-4}
\cmidrule(rl){5-7}
\cmidrule(rl){8-10}
\cmidrule(rl){11-13}
&  R@1 & R@ 2 & R@3 & R@1 & R@ 2 & R@3 & R@1 & R@ 2 & R@3 & R@1 & R@ 2 & R@3 & Average R@1 \\
\midrule
\color{Gray}{Specific Solution (Focus Attribute)} &  \color{Gray}{20.8} & \color{Gray}{32.6} & \color{Gray}{41.1} & \color{Gray}{-} & \color{Gray}{-} & \color{Gray}{-} & \color{Gray}{-}       & \color{Gray}{-}       & \color{Gray}{-} & \color{Gray}{-}       & \color{Gray}{-}       & \color{Gray}{-} & \color{Gray}{-}       \\
\color{Gray}{Specific Solution (Change Attribute)} &  \color{Gray}{-} & \color{Gray}{-} & \color{Gray}{-} & \color{Gray}{15.2} & \color{Gray}{25.8} & \color{Gray}{35.6} & \color{Gray}{-}       & \color{Gray}{-}       & \color{Gray}{-} & \color{Gray}{-}       & \color{Gray}{-}       & \color{Gray}{-} & \color{Gray}{-}       \\
\color{Gray}{Specific Solution (Object)} &  \color{Gray}{-} & \color{Gray}{-} & \color{Gray}{-} & \color{Gray}{-} & \color{Gray}{-} & \color{Gray}{-} & \color{Gray}{18.7}       & \color{Gray}{30.3}       & \color{Gray}{37.4} & \color{Gray}{18.1}       & \color{Gray}{28.7}       & \color{Gray}{34.5} & \color{Gray}{-}       \\
\midrule
Image Only        &  17.7       & 30.9       & \textbf{41.9} & 11.9       & 20.8       & 28.8 & 9.3       & 18.2       & 26.2 & 7.2       & 16.7       & 24.9   & 11.5    \\
Text Only         &  10.2       & 20.5       & 29.6 & 9.5       & 17.6       & 26.4 & 6.5       & 16.8       & 22.4 & 6.2       & 13.9       & 21.4    & 8.1   \\
Image + Text      &  15.6       & 26.3       & 37.1 & 12.6       & 22.9       & 32.0 & 10.8       & 21.0       & 31.2 & 11.3       & 21.5       & 30.3   & 12.6    \\
Combiner (CIRR)      &  15.1       & 27.7       & 39.8 & 12.1       & 22.8       & 31.8 & 13.5       & 25.4       & \textbf{36.7} &     15.4   & 28.0       & 39.6   & 14.0    \\
\midrule
Combiner (CC3M, Ours)              &  \textbf{19.0}       & \textbf{31.0}       & 41.5 & \textbf{16.6}       & \textbf{27.5}       & \textbf{36.5} & \textbf{14.7}       & \textbf{25.9}       & 36.1 & \textbf{16.8}       & \textbf{29.1}       & \textbf{39.7}  & \textbf{16.8}    \\   
\bottomrule
\end{tabular}
}
\vspace{-5mm}
\end{table*}

We evaluate baselines, task-specific solutions, and our method on the proposed GeneCIS benchmark.
~\cref{sec:baselines_and_specific_solutions} describes the baselines as well as specific solutions which we design for each of the GeneCIS tasks. 
~\cref{sec:main_results} shows results on GeneCIS and, in ~\cref{sec:prior_work}, 
we evaluate on related benchmarks from the Composed Image Retrieval (CIR) literature.

\subsection{Baselines and Specific Solutions for GeneCIS}
\label{sec:baselines_and_specific_solutions}

\par \noindent \textbf{CLIP-Only Baselines:}
We provide three simple CLIP-only \cite{Radford2021Learning} baselines for GeneCIS.
Our \textbf{Image Only} baseline embeds all images with the CLIP image encoder and retrieves the closest gallery image to the reference.
The \textbf{Text Only} baseline embeds the text condition with the CLIP text encoder, and the gallery images with the image encoder, and finds the closest gallery image to the text embedding.
Finally, our \textbf{Image + Text} baseline averages the reference image with the condition text feature, before using the combined vector to find the closest gallery image.

\par \noindent \textbf{CIRR Combiner baseline:}
CIRR is a natural image dataset \cite{liu2021cirplant} containing $28$K curated retrieval templates. 
All templates contain a human-specified text condition defining the relationship between the reference and `positive' target image.
Unlike our automatic and scalable triplet mining method, CIRR is manually constructed with a lengthy annotation process.
We include a baseline from \cite{baldrati2022conditioned}, which trains a Combiner model with a CLIP backbone on CIRR.
For fair comparison with our method, we fine-tune both the image and text backbones on CIRR before evaluating the model zero-shot on GeneCIS, terming it \textbf{Combiner (CIRR)}.

\par \noindent \textbf{Specific Solutions:}
We also design specific solutions for each of the proposed tasks in \benchmark. 
These solutions take into account the construction mechanisms of each task and represent sensible approaches to tackling the tasks independently.
We design all solutions to respect the zero-shot nature of the evaluations and hence they are all based on `open-vocabulary' models;
we use CLIP for the attribute-based tasks and Detic \cite{zhou2022detecting} for the object-based ones.
For the attribute-based tasks, we use CLIP to predict attributes or categories in the reference image, before using text embeddings of these predictions to search the gallery. 
For the object-based tasks, we use Detic to detect the object categories present in all images, treating the detected categories as bag-of-word descriptors of the target images. 
We give full details of the specific solutions in \cref{sec:supp_specific_solutions}.

\subsection{Implementation Details}
We train our strongest model on 1.6M triplets mined from Conceptual Captions 3 Million (CC3M) \cite{sharma2018conceptual} which contains 3M image-caption pairs.
Each triplet has a visual concreteness of at least 4.8 averaged over the `Subject' and `Object' entities in both the reference and target image. 
We train the contrastive loss with temperature $\tau\!=\!0.01$ and batch size of 256, training for $28$K gradient steps. 
We use early stopping based on the Recall@1 on the CIRR validation set and, for fair comparison with \cite{baldrati2022conditioned}, initialize the image and text backbones with the ResNet50$\times 4$ CLIP model.
Further details are in \cref{sec:supp_implementation}.

\subsection{Analysis on GeneCIS}
\label{sec:main_results}

We report results for all methods on the GeneCIS benchmark in ~\cref{tab:main}. 
Our evaluation metric is Recall@$K$: the frequency with which the model ranks the `correct' gallery image in its top-$K$ predictions.
We report results at $K=\{1, 2, 3\}$ to evaluate under different constraints, and to account for any noise in the benchmark.
We also report the Average R@1 over all tasks to measure the overall performance across different forms of conditional similarity.

\par \noindent \textbf{Takeaways:}
From the \textit{baselines} we find that both the `Image Only' and `Text Only' models perform poorly as expected, since they only rely on either the reference image content or the text condition.
The `Image + Text' and `Combiner (CIRR)' models perform better, validating our claim that both the reference and text condition are required to solve the task.
Phrased differently, this suggests the benchmark evaluates conditional similarity, as implicit similarity functions (\eg the `Image Only' baseline) perform poorly on average.
We further find that \textit{our method}, using automatically mined data, substantially outperforms all baselines on average across the tasks, as well as at Recall@1 on all tasks individually. 
Notably, it outperforms the model trained on manually collected data from CIRR. 

As expected, most per-task \emph{specific solutions} perform better than our general method.
However, the broad zero-shot nature of GeneCIS makes all tasks independently challenging and the specific solutions do not work for all of them.
Broadly speaking, we found that CLIP \cite{Radford2021Learning} struggles to predict object attributes, and that Detic \cite{zhou2022detecting} struggles on the `stuff' categories in COCO Panoptic \cite{coco2017panoptic}.

Finally, \textit{caveats} can be found in `Image Only' results on `Focus Attribute', where the baseline performs slightly better than our method at higher recalls. 
This is because there are some similarity conditions (\eg `color') for which standard image embeddings are well suited.
We also find that `Combiner (CIRR)' performs better on tasks with object conditions, as the multi-object image distribution of CIRR is more closely aligned with these tasks, than with the single-object images in the attribute-based tasks. 
We note that good performance on all tasks collectively indicates strong general conditional similarity models.

\subsection{Comparisons to Prior Work}
\begin{table}[t]
\footnotesize
\centering
\setlength\belowcaptionskip{0pt}
\caption{\textbf{Results on MIT-States} \cite{StatesAndTransformations}. 
\textit{Zero-shot evaluation} of our model outperforms SoTA supervised methods on this dataset.
}
\label{tab:mit_states}
\resizebox{\linewidth}{!}{
\begin{tabular}{lclll}
\toprule
& Zero-shot & Recall @ 1 & Recall @ 5 & Recall @ 10 \\
\midrule

\color{Gray}{TIRG} \cite{vo2019composing} & \xmark & \color{Gray}{12.2}       & \color{Gray}{31.9}       & \color{Gray}{43.1}       \\
\color{Gray}{ComposeAE} \cite{Anwaar_2021_WACV} & \xmark & \color{Gray}{13.9}       & \color{Gray}{35.3}       & \color{Gray}{47.9}       \\
\color{Gray}{LBF} \cite{hosseinzadeh2020composed} & \xmark & \color{Gray}{14.7}       & \color{Gray}{35.3}       & \color{Gray}{46.6}       \\
\color{Gray}{HCL} \cite{xu2021hierarchical} & \xmark & \color{Gray}{15.2}       & \color{Gray}{36.0}       & \color{Gray}{46.7}       \\
\color{Gray}{MAN} \cite{fu2021multi} & \xmark & \color{Gray}{15.6}       & \color{Gray}{36.7}       & \color{Gray}{47.7}       \\

\hline
Image Only        & \cmark & 3.7       & 14.1       & 22.9       \\
Text Only         & \cmark &  9.5       & 22.5       & 31.4       \\
Image + Text      & \cmark & 13.3       & 31.7       & 42.6       \\
\hline
Combiner (CC3M, Ours)              & \cmark & \textbf{15.8}       & \textbf{37.5}       & \textbf{49.4}      \\   
\bottomrule
\end{tabular}
}
\vspace{-3mm}
\end{table}
\begin{table}[t]
\footnotesize
\centering
\setlength\belowcaptionskip{0pt}
\caption{\textbf{Results on CIRR} \cite{liu2021cirplant}.
Our model substantially outperforms the comparable zero-shot baselines.
}
\label{tab:cirr}
\resizebox{\linewidth}{!}{
\begin{tabular}{lclll}
\toprule
& Zero-shot & Recall @ 1 & Recall @ 5 & Recall @ 10 \\
\midrule

\color{Gray}{ARTEMIS} \cite{delmas2022artemis} & \xmark & \color{Gray}{17.0}       & \color{Gray}{46.1}       & \color{Gray}{61.3}       \\
\color{Gray}{CIRPLANT} \cite{liu2021cirplant} & \xmark & \color{Gray}{19.6}       & \color{Gray}{52.6}       & \color{Gray}{68.4}       \\
\color{Gray}{Combiner (CIRR, \cite{baldrati2022conditioned})} & \xmark & \color{Gray}{38.5}       & \color{Gray}{70.0}       & \color{Gray}{81.9}       \\
\color{Gray}{Combiner (CIRR, improved)} & \xmark & \color{Gray}{40.9}       & \color{Gray}{73.4}       & \color{Gray}{84.8}       \\
\hline
Image Only        & \cmark & 7.5      & 23.9      & 34.7       \\
Text Only         & \cmark & 20.7       & 43.9       & 56.1       \\
Image + Text      & \cmark & 21.8      & 50.9       & 63.7      \\
\hline
Combiner (CC3M, Ours)              & \cmark & \textbf{27.3}       & \textbf{57.0}       & \textbf{71.1}      \\   
\bottomrule
\end{tabular}
}
\vspace{-4mm}
\end{table}
\label{sec:prior_work}

GeneCIS uses natural images with general conditions, rather than being specialized to domains such as bird species \cite{mishraPAN2021}, faces \cite{Zhong16} or fashion compatability \cite{vasileva2018learning,han2017learning,guo2019fashion,han2017automatic}.
As such, to find comparable existing benchmarks, we turn to the \textit{Composed Image Retrieval} (CIR) literature.
The CIR task is to retrieve images which best match a composed reference image and editing text condition.
This task aligns with the `Change' dimension of GeneCIS.
We evaluate on both the MIT-States benchmark \cite{StatesAndTransformations} as well as on CIRR \cite{liu2021cirplant}, with the former precisely reflecting the `Change Attribute' GeneCIS task.

\par \noindent \textbf{Metrics:}
On both benchmarks, we evaluate our model \textit{zero-shot} on the test-sets and compare with prior work trained on the datasets.
These datasets are partially labeled and evaluate using a global retrieval setting, \ie, the entire test-set is used as a gallery for each query.
Thus, we follow prior work and report Recall@K at multiple $K=\{1, 5, 10\}$ to fully capture the model's performance.
\noindent\footnote{CIRR also has an evaluation on curated galleries, akin to GeneCIS. We do not report on this as we found that the `Text Only' baseline performed comparably with SoTA models on this task, achieving over 60\% Recall@1.}

\par \noindent \textbf{Results:} We show results on MIT-States in ~\cref{tab:mit_states}.
Prior work on this benchmark trains models on the dataset from scratch and thus is not zero-shot.
Nonetheless, \textit{zero-shot evaluation} of our model surpasses state-of-the-art on this task.
However, we note that prior methods use smaller models compared to our pre-trained CLIP backbone.

We report on CIRR in ~\cref{tab:cirr}, evaluating through the official test server and again comparing to methods that train for this setting.
We report results for the Combiner method from the paper~\cite{baldrati2022conditioned} as well as our improved implementation (see ~\cref{sec:baselines_and_specific_solutions}), which are both trained on CIRR.
Our improved implementation is a strong upper bound, surpassing previous fully supervised models.
On zero-shot evaluation, our method surpasses the comparable baselines by a signficant margin across all the recall metrics.
Compared to supervised methods, our model outperforms \cite{delmas2022artemis} and \cite{liu2021cirplant} zero-shot, though we note \cite{delmas2022artemis} trains from scratch.
Finally, our model reduces the gap between the baselines and specialist Combiner models trained on CIRR.

\vspace{-2mm}
\section{Analysis}
\par \noindent \textbf{Ablations:} 
~\cref{tab:ablation} shows the effect of our design choices on the performance on GeneCIS.
We find that filtering out relationships which are not visually concrete, and finetuning the entire backbone, both strongly affect the performance.
We verify the robustness of our triplet mining procedure by training with SBU Captions \cite{ordonez2011sbu}, a smaller but different source of image-caption data.
We find that though the larger CC3M \cite{sharma2018conceptual} produces slightly better results, different image-caption datasets are also suitable.

\begin{table}[t]
\footnotesize
\centering
\setlength\abovecaptionskip{3pt}
\setlength\belowcaptionskip{8pt}
\caption{\textbf{Ablations} of key design choices of our full model with
results reported on our \benchmark benchmark.
}
\label{tab:ablation}
\resizebox{\linewidth}{!}{
\begin{tabular}{lr}
\toprule
                  & Average Recall @ 1 \\
\midrule
\rowcolor{Gray}
\textbf{Full Model} & \textbf{16.8} \\
\hline
No filtering for visual concreteness         & 15.0     \\
\hline
Freezing CLIP image backbone  & 14.7       \\
Freezing CLIP text backbone         & 15.8      \\
Freezing entire backbone        & 15.1       \\
\hline
Training on SBU \cite{ordonez2011sbu} instead of CC3M \cite{sharma2018conceptual} caption data         & 16.5    \\
\bottomrule
\end{tabular}
}
\vspace{-6mm}
\end{table}
\par
\begin{figure}[t]
    \centering
    \includegraphics[width=\linewidth]{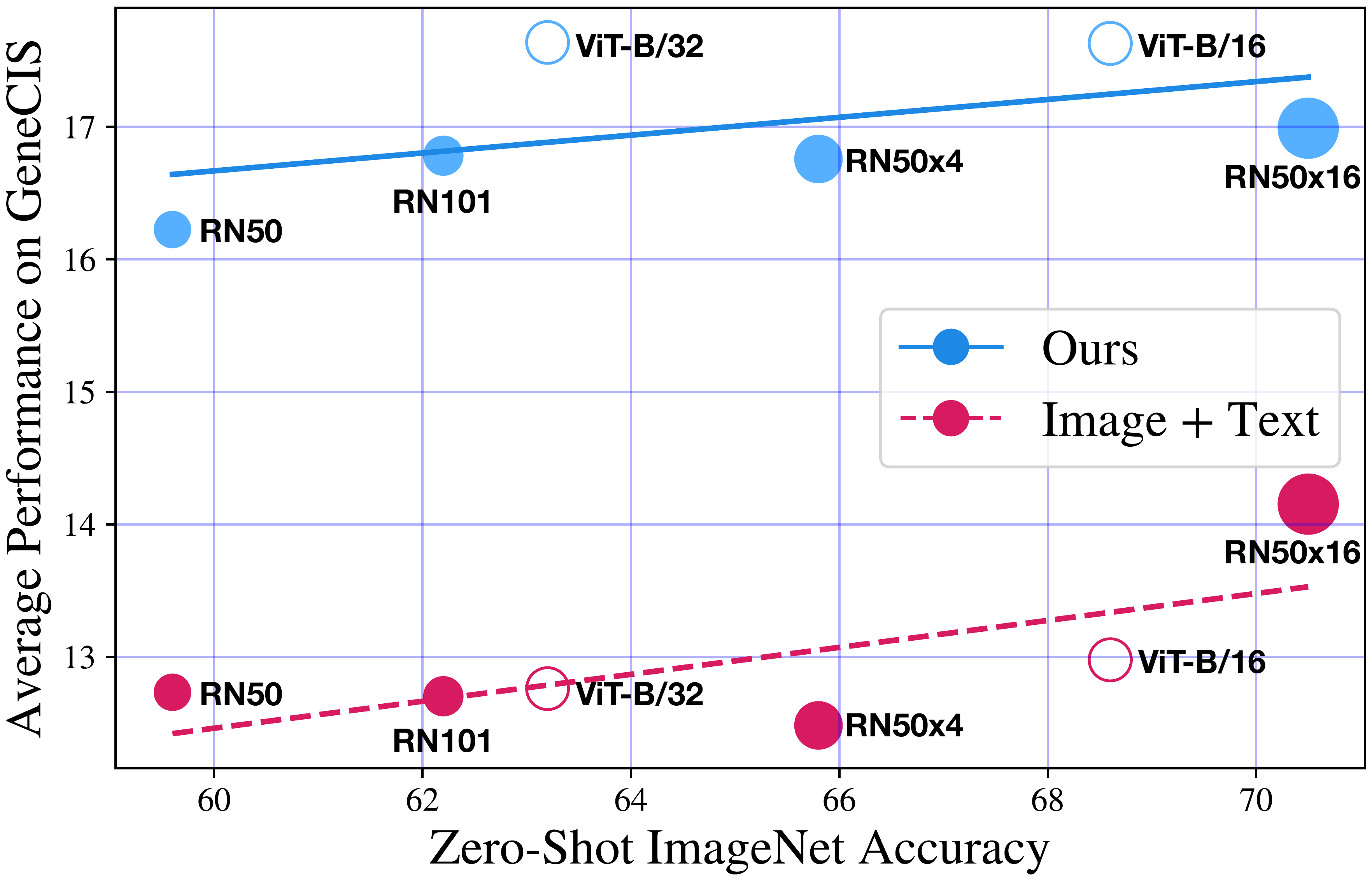}
    \caption{\textbf{Impact of different CLIP backbones} on the performance of our model and the `Image + Text' baseline.
    We show the Average Recall@1 on GeneCIS against the backbones' zero-shot ImageNet accuracy, showing the two have a weak correlation.
    }
    \label{fig:imagenet_correlation}
\end{figure}

 \noindent \textbf{Comparing pretrained backbones:} In ~\cref{fig:imagenet_correlation}, we study the effect of changing the CLIP initialization.
We train Combiner models with ResNet \cite{he2016deep} and ViT \cite{dosovitskiy2020image} backbones on CC3M, showing their performance as well as the `Image + Text' baseline from ~\cref{sec:baselines_and_specific_solutions}.
\footnote{
For fair comparison with \cite{baldrati2022conditioned}, we report with a ResNet50$\times4$ backbone in ~\cref{tab:main}, and report on our strongest ViT-B/16 model in \cref{sec:supp_vit_results}.
}

We plot the performance on GeneCIS against the CLIP backbone's zero-shot ImageNet accuracy \cite{deng2009imagenet}.
We observe that the performance on \benchmark is \textbf{weakly correlated with the ImageNet performance} of the backbone: a Top-1 gain of 10\% on ImageNet leads to only 1\% improvement on \benchmark.
This suggests that improvements on ImageNet do not directly transfer to \benchmark and that \benchmark measures a different yet important capability of vision models.
In addition, our method offers a substantial boost over the `Image + Text' baseline, and a greater boost than scaling the underlying CLIP model.
Both of these results are in stark contrast to trends on popular vision tasks such as segmentation~\cite{li2022language} and detection~\cite{minderer2022simple}, where gains on ImageNet directly transfer to large gains on the downstream task, and often more significantly so than gains from the underlying method.

\par
\begin{figure}[t]
    \centering
    \includegraphics[width=\linewidth]{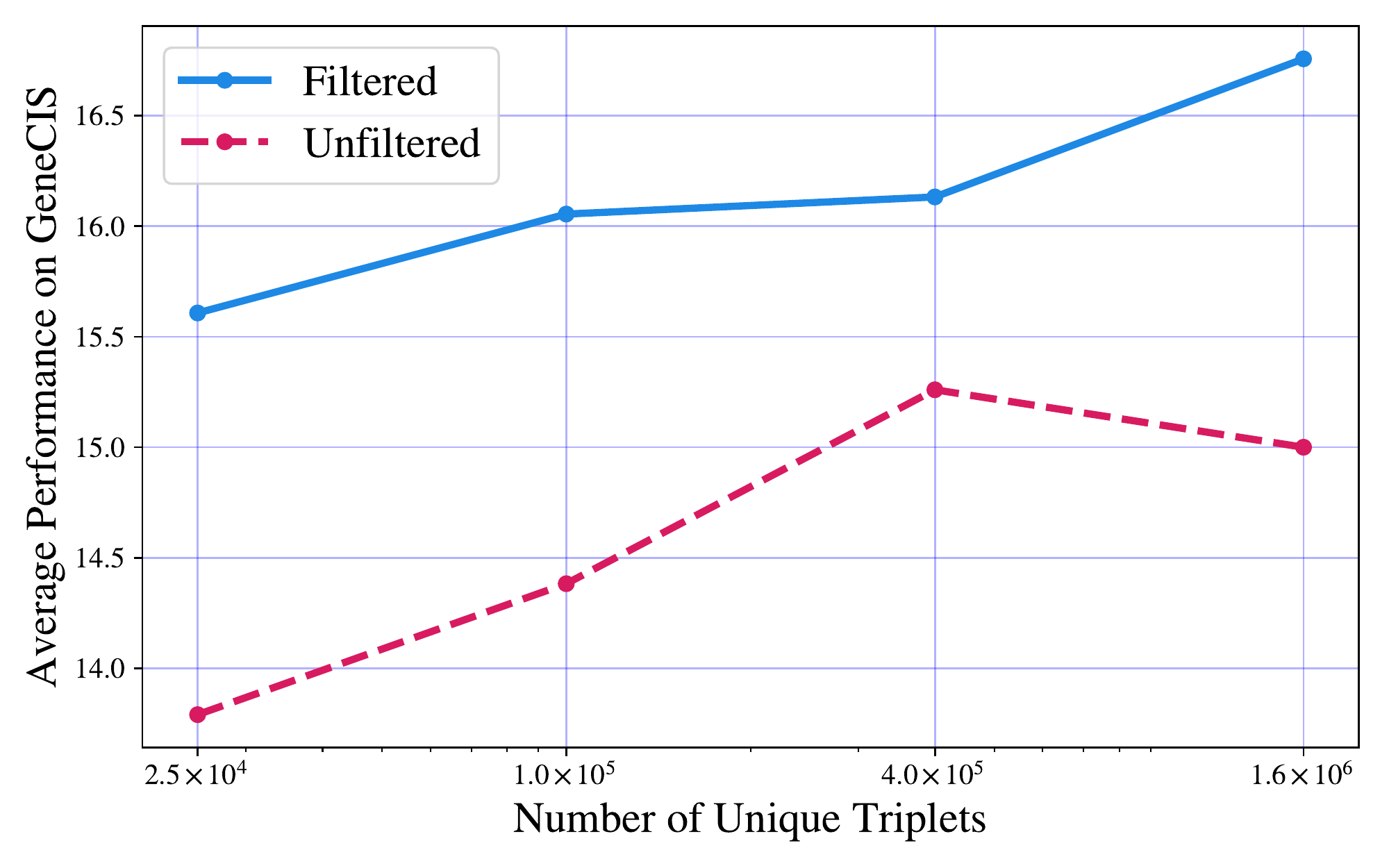}
    \caption{\textbf{Scaling the number of mined triplets} used for training our model improves the performance.
    This suggests that our automatic mining strategy is a promising and scalable approach to learning general similarity functions.
    }
    \label{fig:scaling}
\end{figure}

 \noindent \textbf{Scaling the number of triplets:}
In ~\cref{fig:scaling}, we investigate the effect of scaling the conditional similarity \textit{training data}. 
We successively decrease the number of mined triplets by factors of four (from the $1.6$M used to train our strongest models) both with and without concreteness filtering.
We find results improve with increasing numbers of triplets and that while our models are trained on a dataset of 3M image-caption pairs \cite{sharma2018conceptual}, open-source caption datasets exist with up to five billion images \cite{schuhmann2022laion}.
We emphasize the utility of this finding, suggesting it is possible to train stronger conditional similarity models by further scaling the training data. 

\section{Conclusion}

In this paper we have proposed the GeneCIS benchmark for General Conditional Image Similarity, an important but understudied problem in computer vision.
The benchmark extends prior work and evaluates an open-set of similarity conditions, by being designed for zero-shot testing only.
Furthermore, we propose a way forward for scalably training conditional similarity models, which mines information from widely available image-caption datasets.

Our method not only boosts performance over all baselines on GeneCIS, but also provides substantial zero-shot gains on related image retrieval tasks.
Moreover, we find that unlike for many popular vision tasks, the performance of our models on GeneCIS is roughly decorrelated from scaling the backbone network's ImageNet accuracy, motivating further study of the conditional similarity problem. 

{\small
\bibliographystyle{ieee_fullname}
\bibliography{egbib}
}

\clearpage
\appendix

\addcontentsline{toc}{section}{Appendix} %
\part{Appendix} %
\parttoc %

\par\noindent We provide additional details and discussion of components of the main paper.
We particularly highlight \cref{sec:supp_genecis_details} for details on GeneCIS construction, and \cref{sec:qual_results} for qualitative examples.

\section{Further GeneCIS Details}
\label{sec:supp_genecis_details}

Here, we provide details on the construction process of each GeneCIS task, making reference to the examples from \cref{fig:setting} for clarity.

\subsection{Task construction}
\label{sec:supp_task_construction}

\par \noindent \textbf{Focus on an Attribute:}
VAW \cite{Pham_2021_CVPR} contains bounding box annotations for various objects, as well as a list of \textit{positively labelled attributes} and \textit{negatively labelled attributes} for each object.
Note that, as discussed in \cref{sec:condsim_challenges}, it is impossible to exhaustively label an object for all possible `positive' attributes.
It is, however, possible to determine a set of `negative' attributes. 
For instance, one \textit{cannot} exhaustively label a `thick' tree trunk as \{`wide', `fat', `large'...\etc\}, but one \textit{can} determine that it is not `thin'.

For this task, we construct templates by first sampling a reference object (e.g `laptop') and identifying all \textit{positive attributes} of the object (\eg `white', `plastic') and their corresponding \textit{attribute type} (\eg `color', `material'). 
Given an attribute type, we select a `correct' target image to have the same object category and attribute within the attribute type as the reference (a `laptop' with the same `color'). 
Distractors are then mined to have the same object category but to be explicitly \textit{negatively labelled} for the reference attribute (\eg laptops which are negatively labelled for `white').
The condition in this case is the attribute type (`color'). 

\par \noindent \textbf{Change an Attribute:}
We first select an anchor \textit{attribute type} (e.g `color'), before choosing a reference image and a `correct' target image which share the same object category, but have different attributes within the attribute type. 
In \cref{fig:setting}, the reference and `correct' target are both have the same object category (`train') but have different `colors'. 
The attribute of the `correct' target is given as the condition (`olive green'), and a model must understand the category of the reference image, as well as the attribute specified in the condition, to solve the problem. 

We include two forms of `distractors' in the gallery.
The first form includes images with the conditioning attribute (`olive green'), but with a different object category (\eg `tent').
These images behave as distractors for models which retrieve based only on the the condition (we include 9 such images). 
We also include 5 images with the reference object category but without the conditioning attribute (e.g `trains' which are `red'), behaving as distractors for models which only use the reference image content.

\par \noindent \textbf{Focus on an Object:}
For tasks where the condition contains an object, we take images of cluttered scenes from the multi-object COCO dataset \cite{lin2014coco}.
We use COCO Panoptic Segmentation \cite{coco2017panoptic} data which contains dense category labels for every pixel in the image.

We first select a reference image and identify all of its constituent object categories, ensuring at least 10 categories are present. 
Next, we construct a set of all images in the dataset with at least 6 objects in common with the reference -- but do not contain \textit{all} reference categories -- and rank them based on the extent of their category intersection ($\mathcal{I}_{Close}$).
We also construct a set of images with very \textit{few} intersecting objects as $\mathcal{I}_{Far}$. 
We consider the set of object category IDs in an image as a `bag-of-words' descriptor for the image scene, with images in $\mathcal{I}_{Close}$ containing a `similar scene' to the reference, and $\mathcal{I}_{Far}$ representing a `different scene'.

We randomly select the `correct' target image from $\mathcal{I}_{Close}$, and the \textit{conditioning object} is selected as one of this image's intersecting objects with the reference (\eg `refrigerator' in \cref{fig:setting}).
The first form of distractors is mined by taking images in $\mathcal{I}_{Close}$ which \textit{do not} have the conditioning object.
These examples confuse models which only use the reference image (there are 9 of these).
Another type of distractor is constructed by taking images from $\mathcal{I}_{Far}$ which \textit{do} have the conditioning object, confusing models which only consider the text condition (there are 5 of these).
In this way, only the target image has both a \textit{similar scene} and also the \textit{conditioning object}, and is thus \textit{conditionally} the most similar image in the gallery. 
In \cref{fig:setting}, only the target image contains a `refrigerator' and is `outside'.
We note that solutions which only match `bag-of-objects' descriptors fail here (\eg those which simply detect all objects in the images): 
the `correct' gallery image is randomly selected from $\mathcal{I}_{Close}$ and does not necessarily contain the highest object category overlap.

\par \noindent \textbf{Change an Object:}
This task is constructed in a similar form to `Focus on an Object', in that we first select a reference image and construct $\mathcal{I}_{Close}$ and $\mathcal{I}_{Far}$.
Differently, in this case, we first select the `correct' gallery image from $\mathcal{I}_{Close}$ as the most similar image which does not have perfect object overlap. 
Next, the conditioning object is selected randomly from the objects which \textit{do} appear in the `correct' gallery image, but \textit{not} in the reference (`ceiling' in the example in \cref{fig:setting}). 
Distractors are constructed from both $\mathcal{I}_{Far}$ and $\mathcal{I}_{Close}$, such that they do, and do not, contain the conditioning object respectively.
There are 5 distractors from $\mathcal{I}_{Far}$ and 9 distractors from $\mathcal{I}_{Close}$.

\subsection{Implementation Details}
\label{sec:supp_genecis_implementation}

\par\noindent\textbf{Attribute-based Tasks:}
A taxonomy of \textit{attribute types} is provided in VAW \cite{Pham_2021_CVPR}, containing diverse attribute types from `letter color' to `texture'.
We manually clean and refine the taxonomy for our purposes, for instance reassigning many attributes which were assigned to the `other' attribute type. 
The resulting taxonomy contains 45 attribute types with 663 constituent attributes.
We build the tasks such that they are roughly balanced with respect to attribute type, noting that for some attributes it was not possible to construct a suitable retrieval template.
For the `Focus on an Attribute' task, we manually filter attribute types which do not form clear and visually grounded attribute categories. 
Specifically, we filter: \textit{`opinion'}; \textit{`other after'}; \textit{`other physical quality'}; \textit{`state'}; and \textit{`type'}.

Finally, when cropping an object with a bounding box, we dilate the box by a factor of 0.7 in height and width, before padding the resultant image to square with zeroes.
This allows some context to identify the object (we often found it difficult to categorize the image without this), and also maintains the aspect ratio of the underlying object.
We chose the dilation factor which maximized the discrepancy between the `Image + Text' and `Image Only' Recall@1.

\par\noindent\textbf{Object-based Tasks:}
The object-based datasets are derived from the validation set of COCO Panoptic \cite{coco2017panoptic}, containing 57K images with 133 categories.
The categories include `thing' classes like `zebra' and `bench', as well as `stuff' categories like `sand' and `roof'.
We only consider an object category to be present in an image if it occupies more than 1\% of the image pixels.
After a conditioning object is selected, the COCO category name is given as a condition, and we strip miscellaneous identifiers such as `-stuff' and `-other' from the category names.

\subsection{Dataset noise}

Our tasks are built upon manual annotations in the VAW \cite{Pham_2021_CVPR}, COCO \cite{lin2014coco} and Visual Genome \cite{krishnavisualgenome} datasets. 
These are widely used datasets in the vision community and, as such, our tasks should be error free in principle. 
However, we find some templates provide ill-posed problems through noise and ambiguities in the underlying annotations, as well as through bounding box dilation for attribute-based tasks.

Noise in the datasets is easy to understand, constituting instances where an object category or attribute is obviously mislabelled. 
However, the ambiguities are more subtle, and are artefacts of the underlying taxonomies of the datasets. 
For example, in some COCO images, `ceiling lights' are labelled as `ceiling' instead of `light'.
This is not necessarily wrong, but reflects the fact that labels are defined in a one-hot manner and these pixels could refer to either object category.
This is particularly difficult for attribute-based annotations, as interpretations of attributes are highly subjective (\eg the definitions of `wide' and `narrow' are open to interpretation). 
We highlight that such label ambiguity, though underexplored, is present in almost all computer vision datasets, including in ImageNet \cite{vasudevan2022bagel}.

We address this in a number of ways.
Firstly, we ran a version of our method with 10 random seeds as well as on 10 cross-validation splits, finding the standard deviation in Recall@1 on each task to be around 0.2\%.
Although this does not quantify noise in the dataset, it gives an indication of what can be considered `signal' on the tasks.
Secondly, we find that the `Image + Text' baseline outperforms the `Image Only' and `Text Only' baselines on most tasks, suggesting that the tasks measure conditional similarity.
We discuss the exceptional case of `Focus on an Attribute' in \cref{sec:supp_vit_results}.
Thirdly, we evaluate at Recall@$\{1,2,3\}$, to account for any templates in which a `distractor' image (\ie `incorrect' target) in the gallery actually constitutes a valid solution to the problem.
Finally, we are in the process of manually filtering and verifying the templates, presenting the current version as `GeneCIS v0'.

\subsection{Discussion on symmetry}

We highlight that `similarity', as discussed in this paper, does not describe a \textit{symmetric} mathematical property.
In GeneCIS, while the reference image is considered `similar' to the correct target image given the condition, the reverse may not be true.
For instance, in the `Change an Attribute' example in \cref{fig:setting}, the `green train' in the reference is conditionally similar to the `olive green train' target image, given the condition `olive green'. 
However, this target image is \textit{not} similar to the reference image given the same condition.
In general, we find that `Focus' tasks \textit{are} symmetric given the conditions, but `Change' tasks \textit{are not}.
\section{Specific Solutions}
\label{sec:supp_specific_solutions}

We design specific solutions for each of the proposed tasks in \benchmark. 
These solutions take into account the specific construction mechanisms of each task and represent sensible approaches to tackling each task independently.
We design all solutions to respect the `zero-shot' nature of the evaluations and hence they are all based on `open-world' models;
we use CLIP \cite{Radford2021Learning} for the attribute-based tasks and Detic \cite{zhou2022detecting} for the object-based ones.
All descriptions here refer to \cref{fig:setting} for clarity.

\par\noindent\textbf{Focus on an Attribute:} 
Given the attribute type in the condition (e.g `color'), we first task CLIP with predicting the attribute of the reference image.
Specifically, we use the taxonomy of attributes provided in VAW to construct a zero-shot classifier between attributes within that attribute type (\eg \{`red', `blue', `white'\} within `color').
Given the predicted attribute (\eg `white'), we use its text embedding to find the nearest neighbour from the image embeddings of the gallery set.
\par\noindent\textbf{Change an Attribute:} 
We first use CLIP to predict the category of the the reference image, by constructing a zero-shot classifier from the categories in VAW. 
We then compute the text embedding of the concatenated predicted object name and conditioning attribute (\eg `olive green train') and find the nearest neighbour in the gallery.

\par\noindent\textbf{Focus on an Object and Change an Object:}
We use the same specific solution for both of these settings. 
We first use Detic \cite{zhou2022detecting} to detect all object categories in the reference and gallery images, passing it the 2017 COCO Panoptic categories to construct the classifier. 
Next, we filter out any gallery images which \textit{do not} contain the conditioning object category.
Finally, using the detected object categories in a given image, we construct `bag-of-words' descriptors of the reference image and the remaining gallery images.
Specifically, these descriptors are binary vectors for each image (with elements for every COCO Panoptic category) and are set to `1' if a given category is detected in the image.
We use these descriptors to find the most conditionally similar image to the reference from the (filtered) gallery. 

\par\noindent\textbf{Discussion:} 
Note that all of the solutions described here are specialized in two senses.
Firstly, they are designed with the specific task construction method in mind, and hence are not applicable to all tasks as we desire for a general conditional similarity model. 
Secondly, all specific solutions leverage the underlying taxonomy of the \textit{datasets} (VAW \cite{Pham_2021_CVPR} and COCO Panoptic \cite{coco2017panoptic}) used in the benchmark.
\section{Results with ViT-B/16 on GeneCIS}
\label{sec:supp_vit_results}

In \cref{tab:main}, we report results on GeneCIS with a ResNet50$\times$4 backbone for fair comparison with \cite{baldrati2022conditioned}.
However, in \cref{fig:imagenet_correlation}, we demonstrate that our model performs best when intialized with a ViT-B/16 CLIP backbone \cite{Radford2021Learning,dosovitskiy2020image}.
We include results for this model in \cref{tab:supp_vit}, along with the CLIP-only baselines described in \cref{sec:baselines_and_specific_solutions}.

We first note that, with the ViT-B/16 backbone, our model outperforms all CLIP-only baselines, on all tasks and at all recalls. 
Particularly, with the ResNet50$\times$4 backbone in \cref{tab:main}, the `Image Only' baseline outperformed ours on `Focus Attribute' at higher recalls, which is no longer the case here. 
We further note that the `Focus Attribute' task gives anomalous results when comparing the `Image Only' baseline with `Image + Text'.
Specifically, this is the only task for which the `Image + Text' model does not outperform the other baselines. 
On this task, the information given by the condition is an attribute \textit{type}, rather than the attribute itself (\eg `color' rather than `white' in \cref{fig:setting}). 
As such, the condition information likely only confuses existing vision models, and reduces the performance over the `Image Only' baseline.

\begin{table*}[ht]
\footnotesize
\centering
\caption{\textbf{Evaluation on \benchmark with a ViT-B/16 backbone} where we evaluate our method and CLIP-only baselines.
We find our method performs best with a ViT-B/16 backbone and that, with this architecture, our model outperforms the baselines at all recalls on all \benchmark tasks.
}
\label{tab:supp_vit}
\resizebox{\linewidth}{!}{
\begin{tabular}{l|ccc|ccc|ccc|ccc|cc}
\toprule
& \multicolumn{3}{c}{Focus Attribute} & \multicolumn{3}{c}{Change Attribute} & \multicolumn{3}{c}{Focus Object} & \multicolumn{3}{c}{Change Object} \\
\cmidrule(rl){2-4}
\cmidrule(rl){5-7}
\cmidrule(rl){8-10}
\cmidrule(rl){11-13}
&  R@1 & R@ 2 & R@3 & R@1 & R@ 2 & R@3 & R@1 & R@ 2 & R@3 & R@1 & R@ 2 & R@3 & Average R@1 \\
\midrule
Image Only        &  18.1       & 30.1       & 40.6 & 11.5       & 21.9       & 30.9 & 9.4       & 17.0       & 25.4 & 7.6       & 17.1       & 25.5   & 11.7    \\
Text Only         &  10.3       & 20.9       & 30.4 & 10.2       & 18.2       & 26.1 & 7.4       & 14.0       & 23.0 & 8.1       & 16.4       & 24.7    & 
9.0 \\
Image + Text      &  17.1       & 29.5       & 40.5 & 13.1       & 22.2       & 31.9 & 11.5       & 20.1       & 29.2 & 9.8       & 20.0       & 28.9   & 12.9    \\
\midrule
Combiner (CC3M, Ours)              &  \textbf{19.7}       & \textbf{31.7}       & \textbf{42.1} & \textbf{16.2}       & \textbf{27.3}       & \textbf{37.5} & \textbf{16.6}       & \textbf{27.7}       & \textbf{37.2} & \textbf{18.0}       & \textbf{32.2}       & \textbf{41.6}  & \textbf{17.6}    \\   
\bottomrule
\end{tabular}
}
\vspace{-5mm}
\end{table*}
\section{Combiner Architecture}
\label{sec:supp_architecture}

The Combiner architecture takes in reference image and condition text features, composing them into a single vector as:
$g(\mathbf{x}^{R}, \mathbf{e})$ 
where
$g,\mathbf{x}^{R},\mathbf{e} \in  \mathbb{R}^{D}$.

The architecture consists of four functions ($h_i$, built from MLPs) which process features in parallel as:
\vspace{-4mm}

\begin{equation}
    g(\mathbf{x}^{R}, \mathbf{e}) = \lambda  h_1(\mathbf{x}^{R}) + (1 - \lambda) h_2(\mathbf{e}) + h_3(\mathbf{x}^{R}, \mathbf{e})
\end{equation}

where $\lambda = h_4(\mathbf{x}^{R}, \mathbf{e})$ is a dynamic weighting of the features. 
We refer to \cite{baldrati2022conditioned} for full details.
\section{Further Implementation Details}
\label{sec:supp_implementation}

\par\noindent\textbf{Our method:}
All models trained on CC3M were trained for 28K gradient steps.
We train our strongest models with an initial learning rate of 1$\times10^{-6}$ and a cosine decay schedule, training both the CLIP backbone and the Combiner head with the same learning rate.
We evaluate our model at each epoch, selecting the checkpoint with the best Recall@1 on the CIRR validation set \cite{liu2021cirplant}.
Note that this single model is then taken and evaluated \textit{zero-shot} on all benchmarks reported in this paper. 
Our optimizer is Adam \cite{kingma2014adam} and we implement our models using PyTorch \cite{paszke2019pytorch}.
To fine-tune both the backbones and Combiner head with a batch size of 256, we train our models on 16 A100 GPUs, with a training time of approximately 12 hours.
Finally, all features are normalized before the contrastive loss is computed.

\par\noindent\textbf{Specific Solutions:}
For the attribute-based specific solutions, we use the same ResNet50$\times$4 backbone as for our method in \cref{tab:main}. 
Furthermore, when embedding an object name, we ensemble over the 80 standard CLIP prompts from \cite{Radford2021Learning}.
For the object-based baselines, we use a Detic model with the strongest Swin-B backbone \cite{liu2021Swin}, trained for open-vocabulary detection on the `base' classes of the LVIS dataset \cite{gupta2019lvis}. 
For the first detection stage, we select a confidence threshold which optimizes downstream Recall@1 on the `Focus Object' evaluation (a confidence of 0.2).
\section{Extended Related Work}

We briefly describe work from the \textit{text-based image-editing} domain and discuss how it relates to our work. 
Generative image-editing \cite{nichol2021glide, patashnik2021styleclip, brown2022end} is a popular task in which, given a reference image and some condition, a sensible edit of the image is \textit{generated}.
Recently, with the advent of widely available large-scale generative models \cite{rombach2022high}, there has been substantial work which considers the prompt as a text-condition, and uses CLIP text embeddings to guide the generation process \cite{brooks2022instruct,kwon2022clipstyler,kim2022diffusionclip,gal2022stylegan,crowson2022vqgan,avrahami2022blended}.
As such, the inputs and outputs of image-editing models are similar to those considered in this work. 
However, we highlight our work focuses on the \textit{representation} and \textit{retrieval} of existing images, rather than the synthesis of new ones. 

We also note recent work which considers a task similar to `Change an Object' in GeneCIS, in the context of compositional learning \cite{neculai2022probabilistic}. 
Similar to work detailed in \cref{sec:related}, \cite{neculai2022probabilistic} trains on a finite set of categories and considers only one of the four areas of the conditional similarity space which we propose here.
Finally, we highlight \cite{xiao2020should}, which considers contrastively training a single backbone with multiple heads, each of which is invariant to different data augmentations.
This work shares similar motivations to ours -- that there are many notions of image similarity -- though they train with a pre-determined and fixed set of data augmentations (and hence fixed concepts of similarity).

\section{Qualitative Examples}
\label{sec:qual_results}

\subsection{GeneCIS examples}
\label{sec:supp_genecis_examples}

\begin{figure*}[ht!]
    \centering
    \includegraphics[width=0.9\linewidth]{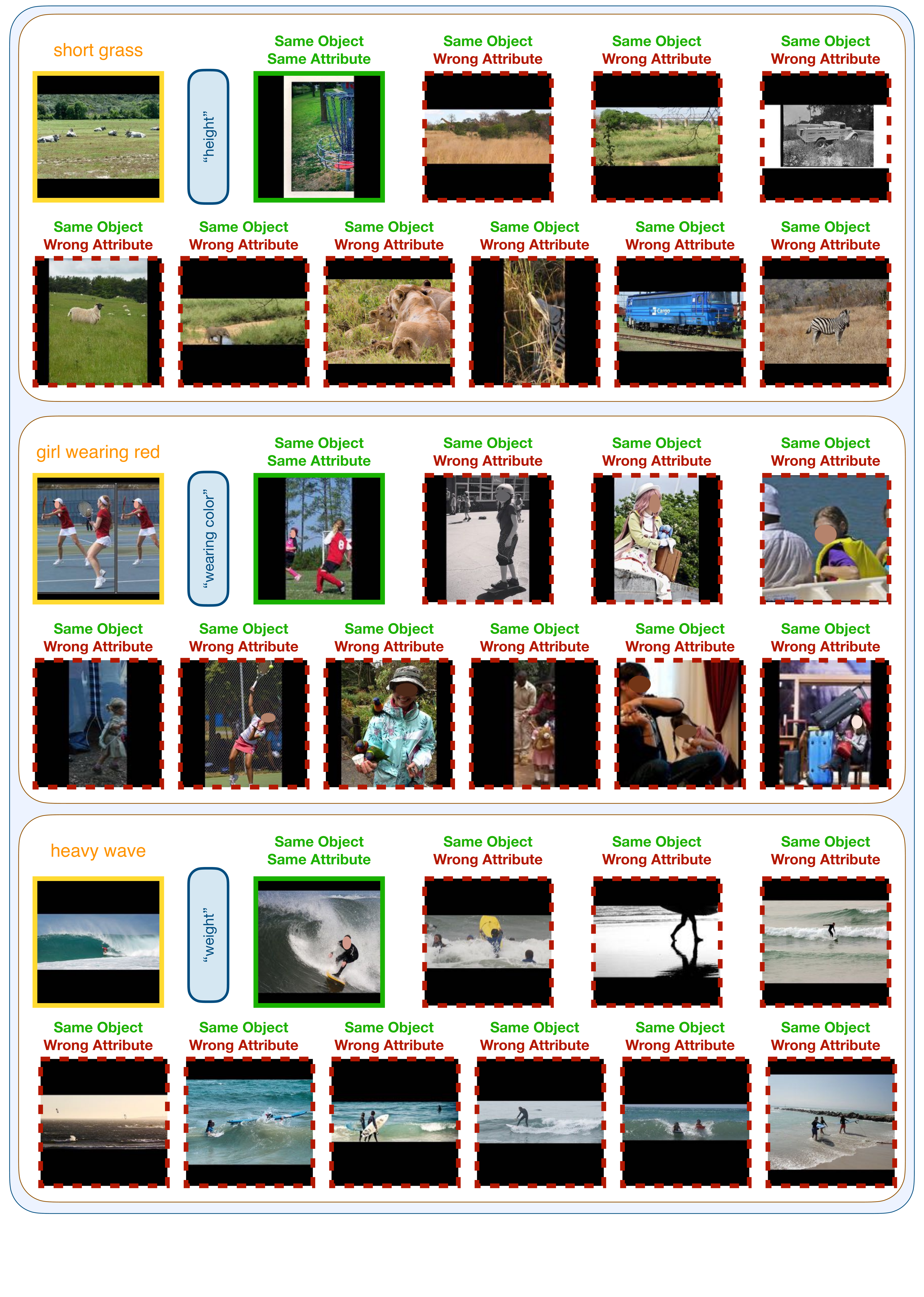}
    \caption{\textbf{Focus on an Attribute example templates.}} 
    \label{sup_fig:focus_attribute}
\end{figure*}

\begin{figure*}[ht!]
    \centering
    \includegraphics[width=0.9\linewidth]{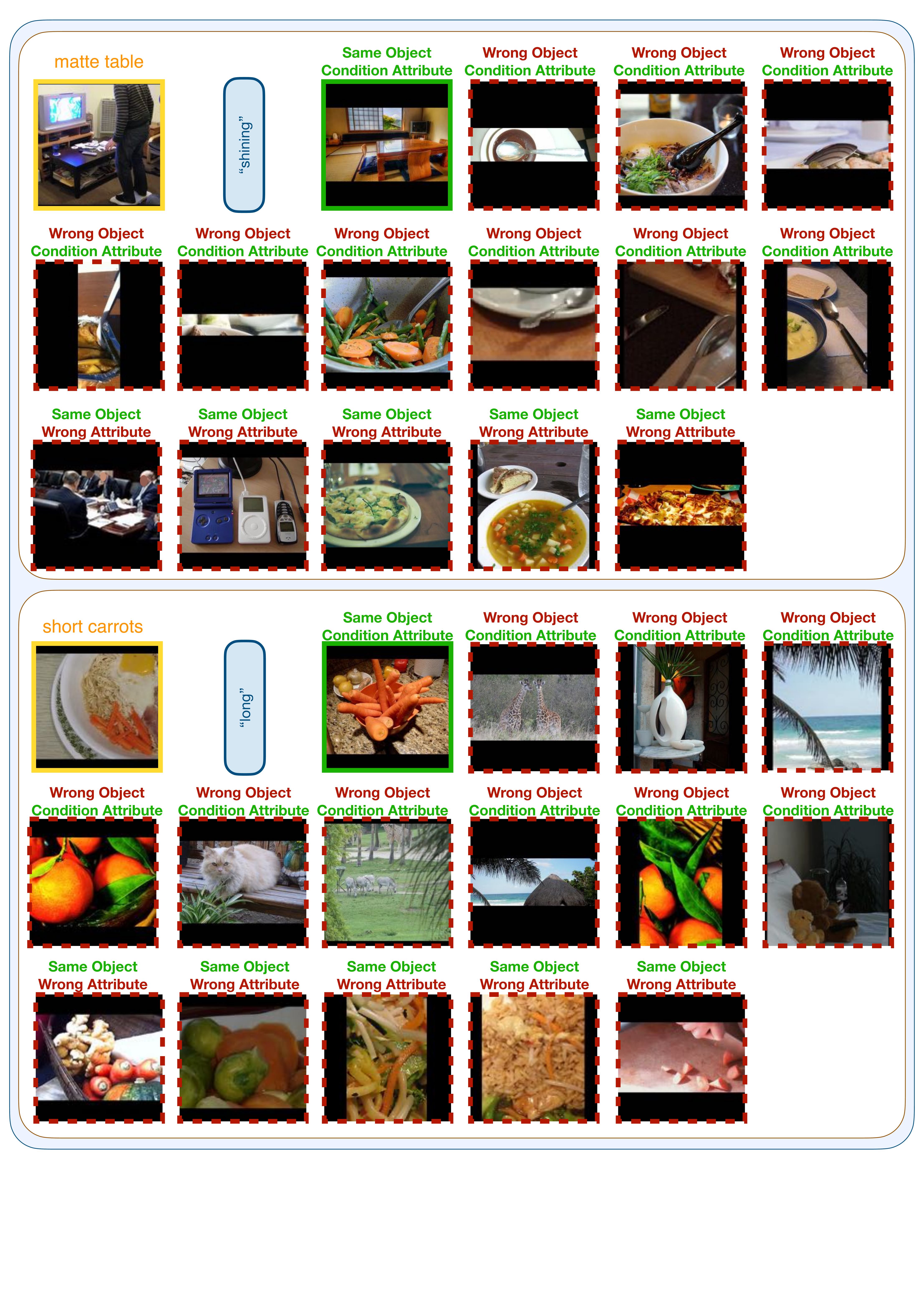}
    \caption{\textbf{Change an Attribute example templates.}} 
    \label{sup_fig:change_attribute}
\end{figure*}

\begin{figure*}[ht!]
    \centering
    \includegraphics[width=0.9\linewidth]{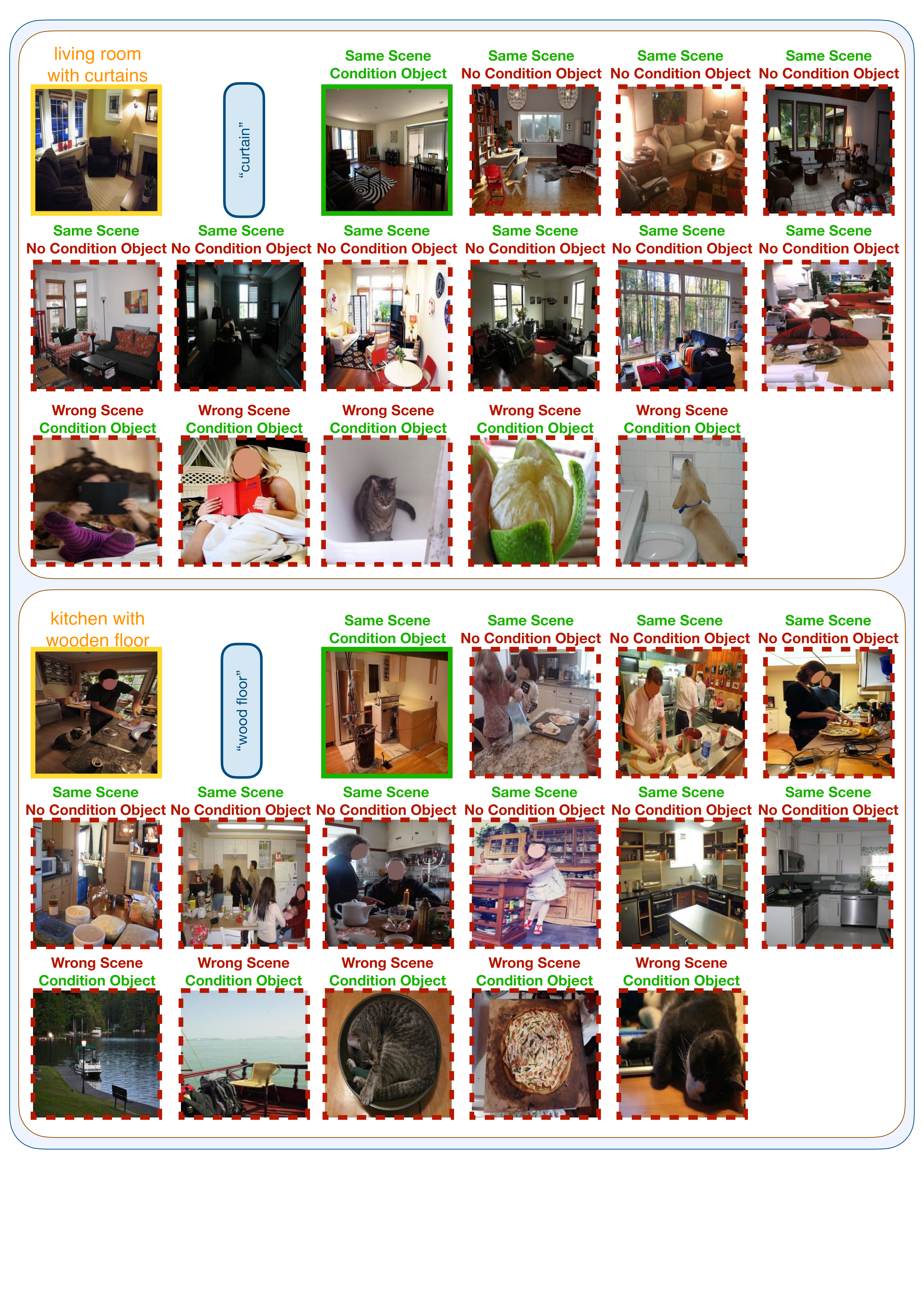}
    \caption{\textbf{Focus on an Object example templates.}} 
    \label{sup_fig:focus_obj}
\end{figure*}

\begin{figure*}[ht!]
    \centering
    \includegraphics[width=0.9\linewidth]{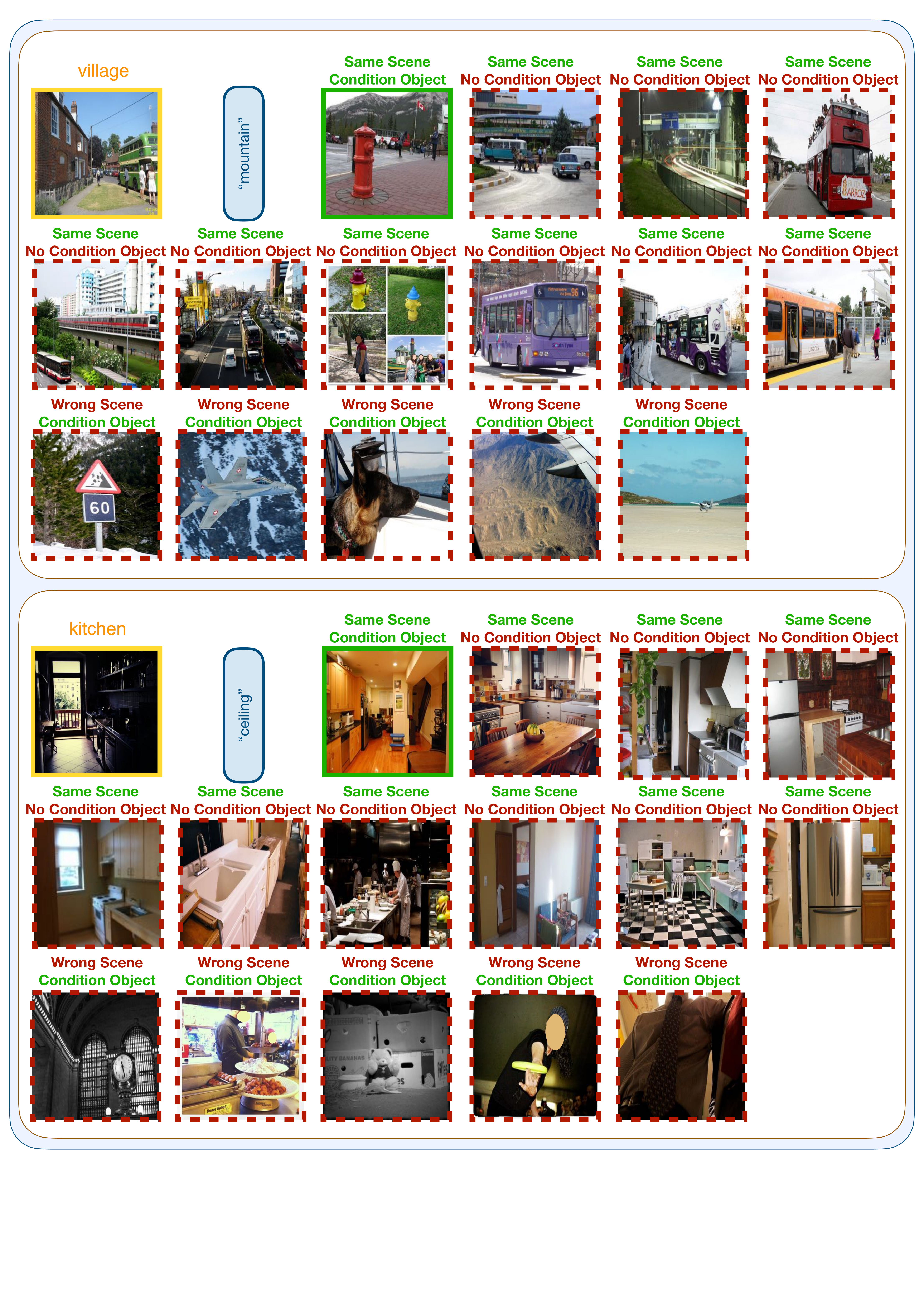}
    \caption{\textbf{Change an Object example templates.}} 
    \label{sup_fig:change_obj}
\end{figure*}

We provide example templates from the GeneCIS benchmark tasks in \cref{sup_fig:focus_attribute,sup_fig:change_attribute,sup_fig:focus_obj,sup_fig:change_obj}.
Differently to \cref{fig:setting}, we show the entire curated retrieval templates of 10-15 target images, as well as the reference image (leftmost, yellow) and condition text (blue oval).
As discussed in \cref{sec:benchmark_details}, all gallery images are \textit{implicitly similar} to the reference image or condition. 
The `positive' target image is the \textit{most similar given the condition}.

\subsection{Model Predictions}

\begin{figure*}[ht!]
    \centering
    \includegraphics[width=0.7\linewidth]{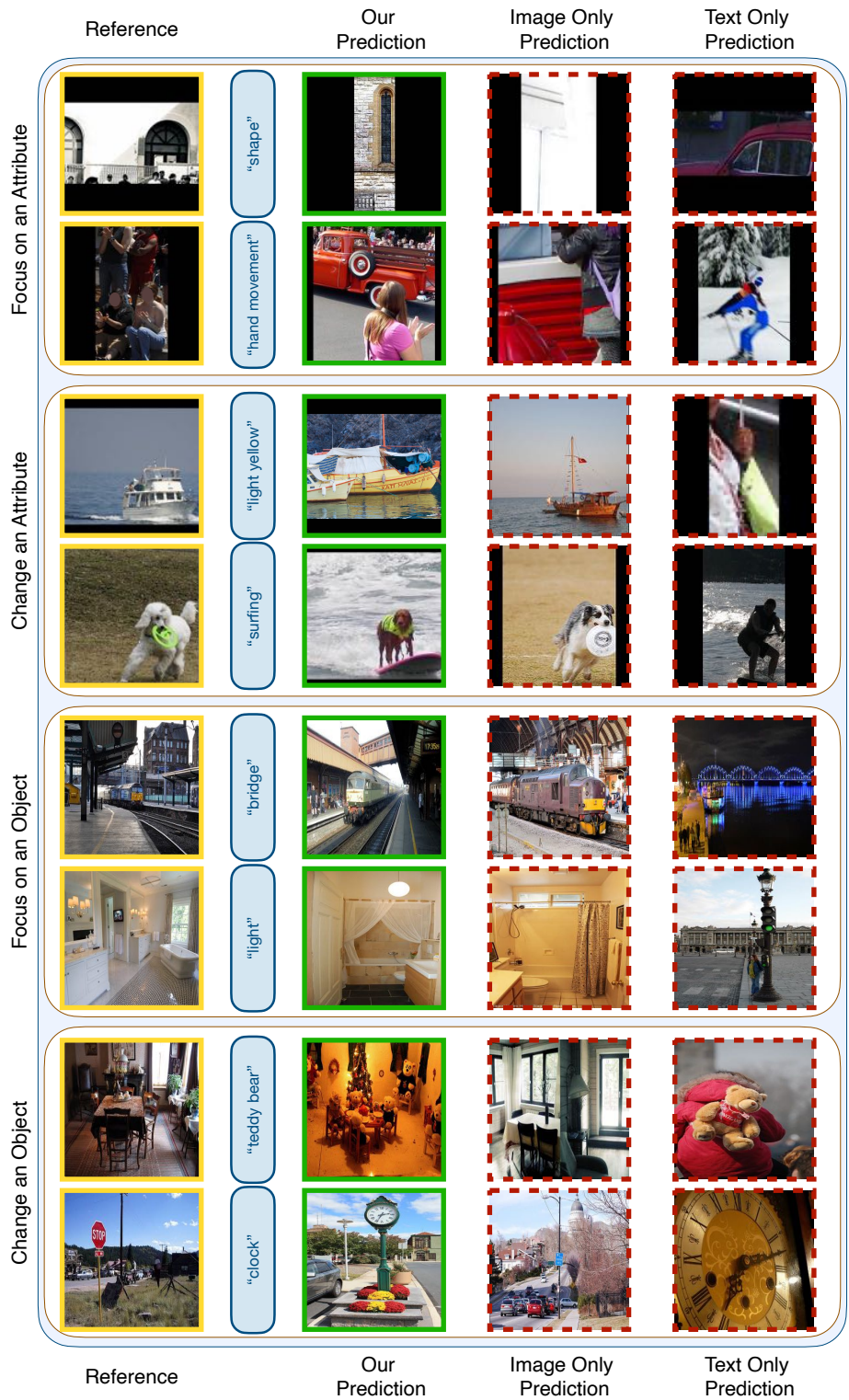}
    \caption{\textbf{Qualitative results from our method} showing instances where our model predicts correctly.}
    \label{fig:supp_preds_qual_good}
\end{figure*}

\begin{figure*}[ht!]
    \centering
    \includegraphics[width=0.7\linewidth]{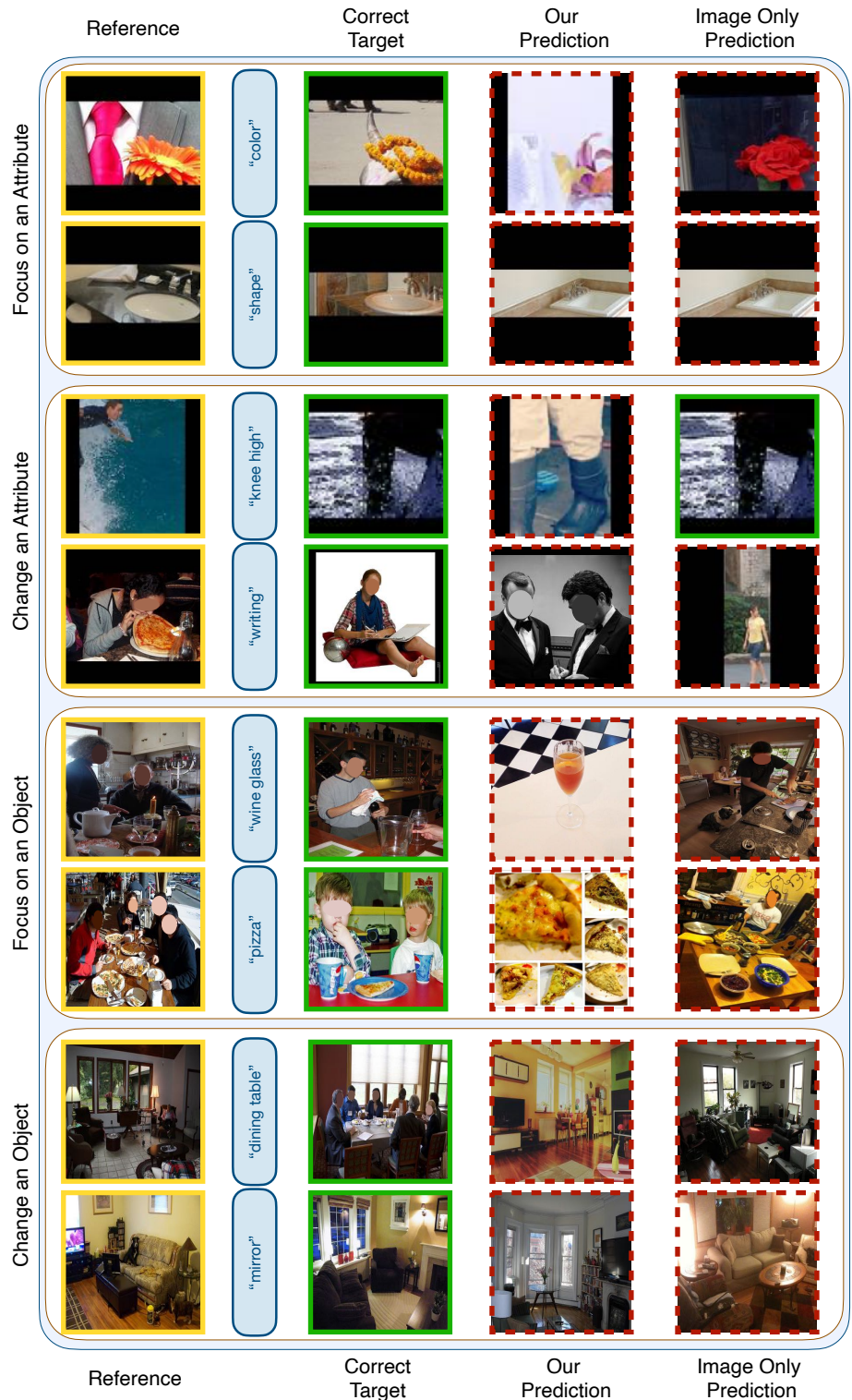}
    \caption{\textbf{Qualitative results from our method} showing instances where our model \textit{fails}.}
    \label{fig:supp_preds_qual_bad}
\end{figure*}

We show qualitative results of our model and CLIP-only baselines in \cref{fig:supp_preds_qual_good}.
We show instances where our model fails in \cref{,fig:supp_preds_qual_bad} along with the `correct' target image and the prediction of the Image-Only CLIP baseline.

\subsection{Mined Triplets from CC3M}

\begin{figure*}[ht!]
    \centering
    \includegraphics[width=0.8\linewidth]{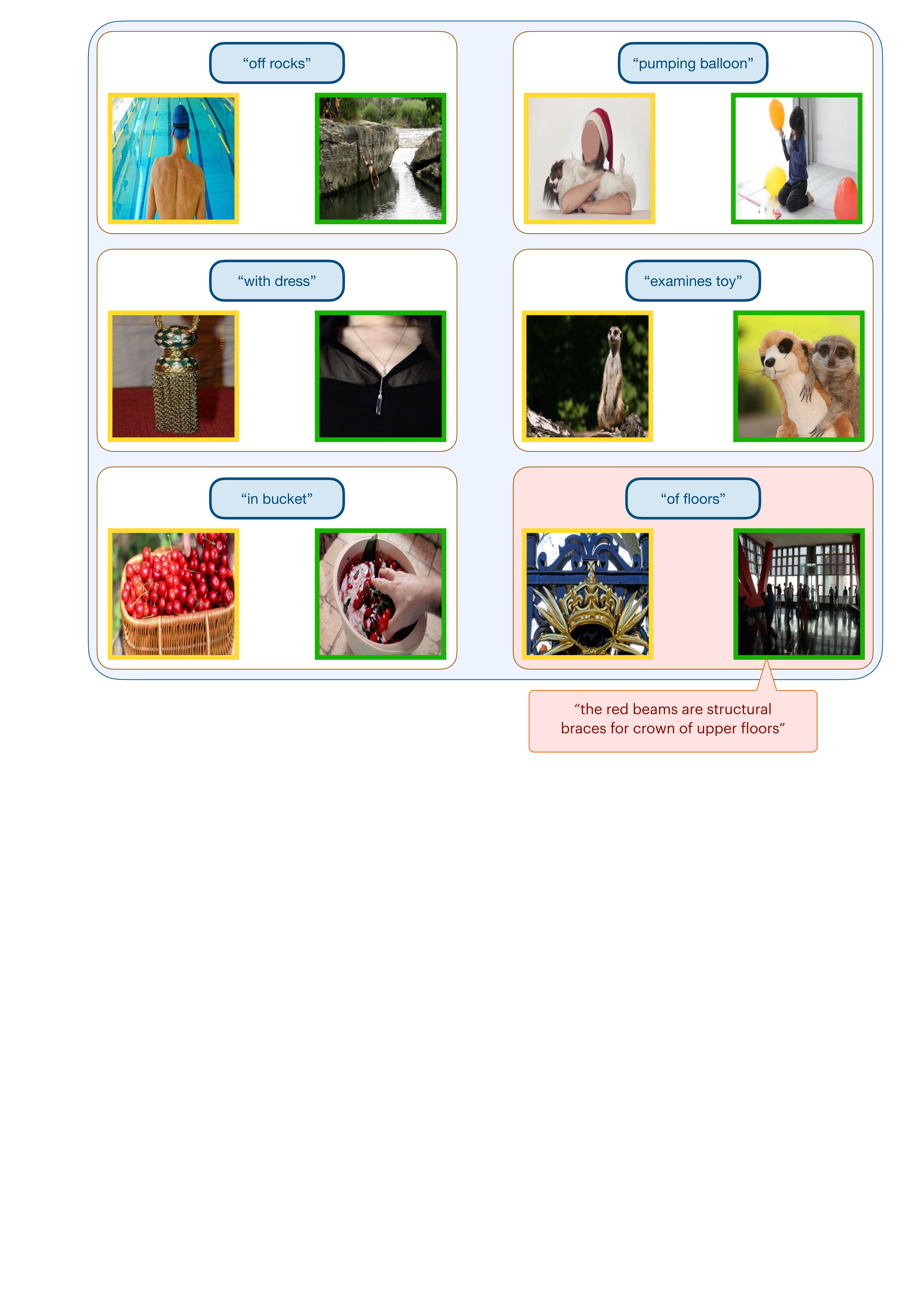}
    \caption{\textbf{Examples of mined triplets from CC3M \cite{sharma2018conceptual}} as described in \cref{sec:triplet_mining}. 
    In each triplet, the text condition (blue oval) links the reference image (left) to the target image (right).
    We show an instance where a noisy triplet is produced in the bottom right. 
    The caption (show in a speech bubble) incurs a misleading parsed relationship of `Crown' $\rightarrow$ `of' $\rightarrow$ `floors'.
    }
    \label{fig:supp_mined_triplets}
\end{figure*}

We show examples of training triplets which we \textit{automatically} mine from CC3M \cite{sharma2018conceptual} in \cref{fig:supp_mined_triplets}.
\section{Attributions}

\cref{fig:motivation}: 
Thomas Hawk, CC BY-NC 2.0 (https://creativecommons.org/licenses/by-nc/2.0/), via Flickr.

\cref{fig:method}, image of horse on canvas:
Simon Kozhin, CC BY-SA 3.0 (https://creativecommons.org/licenses/by-sa/3.0), via Wikimedia Commons.

\section{Acknowledgements}
We would like to thank Weidi Xie, Liliane Momeni, Mannat Singh and Kalyan Vasudev Alwala for valuable discussions on this work. 
Sagar is supported by a Facebook AI Research Scholarship.

\clearpage


\end{document}